\documentclass{article}

\PassOptionsToPackage{round}{natbib}
\usepackage[final,main]{neurips_2026}

\author{
    Bogdan Kulynych \\ Biomedical Data Science Center, Lausanne University Hospital (CHUV) \vspace{-1em} \And
    Shahab Asoodeh \\ McMaster University and Vector Institute
}

\usepackage{nicefrac}
\usepackage{booktabs}
\usepackage{amsmath, amsthm, amssymb}
\usepackage{thmtools, thm-restate}
\usepackage{dsfont}

\usepackage{graphicx}
\usepackage{wrapfig}
\usepackage{multirow}
\usepackage[pdftex,bookmarksnumbered,bookmarksopen,colorlinks,citecolor={blue!70!black},linkcolor={blue!70!black},urlcolor=blue]{hyperref}
\usepackage[capitalize]{cleveref}
\usepackage[ruled]{algorithm2e}
\crefname{algocf}{Algorithm}{Algorithms}
\Crefname{algocf}{Algorithm}{Algorithms}
\crefname{appendix}{Appendix}{Appendices}
\Crefname{appendix}{Appendix}{Appendices}
\usepackage[dvipsnames]{xcolor}
\newcommand{\SA}[1]{}
\newcommand{\BK}[1]{}

\usepackage{pgfplots}
\usetikzlibrary{intersections}
\usetikzlibrary{pgfplots.fillbetween}
\usetikzlibrary{fadings}
\usetikzlibrary{patterns}
\usetikzlibrary{shadows.blur}
\usetikzlibrary{shapes}

\usepackage{tikz}
\usepackage{xcolor}
\usepgfplotslibrary{groupplots,fillbetween}
\usetikzlibrary{arrows.meta,positioning}
\pgfplotsset{compat=1.18}
\usetikzlibrary{calc}

\definecolor{rocketA5}{HTML}{3F1B43}
\definecolor{rocketB5}{HTML}{841E5A}
\definecolor{rocketC5}{HTML}{CB1B4F}
\definecolor{rocketD5}{HTML}{F06043}
\definecolor{rocketE5}{HTML}{F6AB83}
\definecolor{rocketA4}{HTML}{4C1D4B}
\definecolor{rocketB4}{HTML}{A11A5B}
\definecolor{rocketC4}{HTML}{E83F3F}
\definecolor{rocketD4}{HTML}{F69C73}
\definecolor{sbgray}{HTML}{8C8C8C}

\colorlet{ROCColor}{blue!70!black}
\colorlet{TVColor}{black!30}
\colorlet{ChanceColor}{gray}

\title{Integral Probability Boundaries}

\newtheorem{theorem}{Theorem}[section]
\newtheorem{proposition}[theorem]{Proposition}
\newtheorem{definition}[theorem]{Definition}
\newtheorem{lemma}[theorem]{Lemma}
\newtheorem{corollary}[theorem]{Corollary}

\theoremstyle{remark}
\newtheorem{remark}[theorem]{Remark}

\renewcommand{\cite}{\citep}

\newcommand{\indic}{\mathds{1}}
\newcommand{\E}{\mathbb{E}}
\newcommand{\R}{\mathbb{R}}
\newcommand{\F}{\mathcal{F}}
\newcommand{\X}{\mathcal{X}}
\newcommand{\Y}{\mathcal{Y}}
\newcommand{\C}{\mathcal{C}}
\newcommand{\B}{\mathcal{B}}
\newcommand{\define}{~\smash{\triangleq}~}

\begin{document}

\maketitle

\begin{abstract}
\noindent
Bounds based on Integral Probability Metrics (IPMs) are ubiquitous in machine learning and statistics, appearing, e.g., in generalization, alignment, fairness, and privacy. IPMs reduce the discrepancy between two distributions to a single worst-case gap between expectations. This compression is costly: for rare events, safety failures, or under-represented groups, IPM-based bounds are loose or vacuous. We introduce a non-linear generalization of IPMs, which we term Integral Probability Boundaries (IPBs). Unlike IPMs, IPBs provide bounds that are tight at the extremes. We derive their basic properties and present algorithms for estimating them in practically relevant settings. We demonstrate the usage of IPBs in both theoretical and practical applications for alignment, judge-free performance estimation, and fairness, where they enable us to obtain sharp bounds on performance and fairness gaps, e.g., 15 p.p.\@ more precise estimates on fairness gaps for rare groups in a realistic generative modeling task, compared to the state of the art.

\end{abstract}

\section{Introduction}
Bounding expectations of a function over one distribution $P$ in terms of the expectation over another distribution $Q$ is one of the core building blocks in theoretical and applied statistics and machine learning. We often want to analyze or estimate bounds of the following form:
\vspace{-.1em}
\begin{equation}\label{eq:envelope-bound}
    \E_P[f] \leq \E_Q[f] + \delta,
\end{equation}
where the gap $\delta$ could depend on a notion of discrepancy between $P$ and $Q$ such as KL divergence. This form appears across various fields of inquiry, e.g., studies of generalization ($P$ is a test data distribution, $Q$ is a training data distribution), 
alignment ($P$ is an aligned policy, $Q$ is an original policy), fairness ($P$ and $Q$ are data distributions of two population groups), privacy ($P$ and $Q$ are distributions of a model with and without a certain example included in the training data).

\cref{eq:envelope-bound} is also fundamentally related to an object called an \emph{integral probability metric} (IPM)~\cite{muller1997integral} between $P$ and $Q$, which characterizes the maximum difference between the expectations in \cref{eq:envelope-bound} over some family of functions $f \in \F$:
\vspace{-.1em}
\begin{equation}\label{eq:ipm}
    D_\F(P, Q) \define \sup_{f \in \F}|\E_P[f] - \E_Q[f]|.
\end{equation}
It is easy to see that if we can estimate $D_\F(P, Q)$, then we can set $\delta = D_\F(P, Q)$.

Although convenient, bounds as in \cref{eq:envelope-bound} suffer from a critical problem when their sharpness is important. Say that $f(x)$ has a range of $[0, 1]$. Then, at the extremes, when $\E_Q[f] \approx 0$ or $\E_Q[f] \approx 1$, the bound is either overly loose or vacuous. This is because an IPM compresses potentially rich information into a single number. In contrast, consider the entire achievable region of expectations:
\vspace{-.1em}
\begin{equation}
    \C_\F(P ~\|~ Q) \define \{ (\E_Q[f], \E_P[f]) \mid f \in \F\}.
\end{equation}

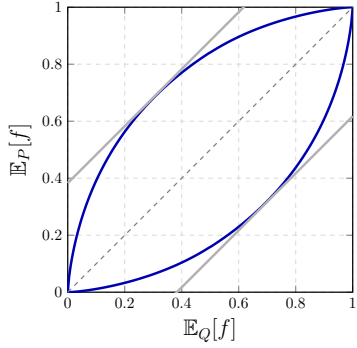
\begin{figure}[t]
    \centering
    \begin{minipage}[c]{0.35\linewidth}
        \centering
        \resizebox{\linewidth}{!}{
        \begin{tikzpicture}[
    declare function={
        normcdf(\x) = 1/(1 + exp(-0.07056*((\x))^3 - 1.5976*(\x)));
        _invpart(\p) = -5.531 * (((1 - \p) / \p)^0.1193 - 1);
        invnormcdf(\p) = (\p >= 0.5) * _invpart(\p) + (\p < 0.5) * (-_invpart(1 - \p));
        g(\x,\mu) = 1 - normcdf(invnormcdf(1 - \x) - \mu);
        g_reflected(\x,\mu) = 1 - g(1 - \x, \mu);
        tv_distance(\mu) = 2 * normcdf(\mu/2) - 1;
    }
]
    \begin{axis}[
        xlabel={\Large $\E_Q[f]$},
        ylabel={\Large $\E_P[f]$},
        domain=0:1,
        xmin=0, xmax=1,
        ymin=0, ymax=1,
        width=8cm,
        height=8cm,
        samples=200,
        axis lines=box,
        axis equal image,
        axis line style={thick},
        enlargelimits=false,
        grid=major,
        grid style={dashed, gray!30},
        legend style={
            at={(0.45,0.05)}, 
            anchor=south west, 
            draw=lightgray, 
            fill=white, 
            fill opacity=0.8, 
            text opacity=1,
            legend cell align={left}
        },
        xtick={0, 0.2, 0.4, 0.6, 0.8, 1},
        ytick={0, 0.2, 0.4, 0.6, 0.8, 1},
        clip=true,
    ]
    
    \def\mu{1}
    \pgfmathsetmacro{\eta}{tv_distance(\mu)}

    \addplot[ROCColor, ultra thick, name path=rocupper] {g(x, \mu)};
    \addplot[TVColor, ultra thick, name path=tvupper, domain=0:1] {x + \eta};
    
    \addplot[ROCColor, ultra thick, name path=roclower] {g_reflected(x, \mu)};
    \addplot[TVColor, ultra thick, name path=tvlower, domain=0:1] {x - \eta};
    
    \addplot[ChanceColor, thick, dashed] {x};

    \tikzfillbetween[of=tvupper and tvlower]{TVColor, opacity=0.1};
    \tikzfillbetween[of=rocupper and roclower]{ROCColor, opacity=0.1};
    
    \end{axis}
\end{tikzpicture}
        }
    \end{minipage}\hfill
    \begin{minipage}[c]{0.55\linewidth}
        \vspace{-2em}
        \caption{\textbf{Integral Probability Boundaries (IPBs) characterize the difference in expectations of any function from a given class $\F$ under two distributions $P$ and $Q$ more precisely than a standard Integral Probability Metric (IPM).} We show the tight IPBs and IPM for $P \sim \mathcal{N}(0, 1)$ and $Q \sim \mathcal{N}(1, 1)$ over the set of all bounded functions
        where the IPM is given by the TV distance $\delta = D_\mathsf{TV}(P, Q)$, and the IPB is given by the ROC of the Neyman--Pearson test (see \cref{sec:properties}). We show IPM-based bounds
        with grey lines (\textcolor{TVColor}{\bf –}), and IPB-based bounds that trace the contour of the achievable region $\C_\F(P ~\|~ Q)$
        with blue lines (\textcolor{ROCColor}{\bf –}).
        }
        \label{fig:ipb}
    \end{minipage}
    \vspace{-1em}
\end{figure}

We introduce a non-linear generalization of IPMs that incorporates a dependency on the exact value of $\E_Q[f]$, and is a boundary of this achievable region:
\begin{definition}[Integral Probability Boundary]\label{def:ipb}
    Consider a pair of probability distributions $(P, Q)$ over a space $\X$, and a family of functions $\F$ with each $f \in \F$ mapping $\X \to \Y \subseteq \R$. Let $\overline{\Y}\define \operatorname{conv}(\Y)$ be the convex hull of $\Y$. A function $g: \overline{\Y} \to \overline{\Y}$ is an integral probability boundary (IPB) of $P$ w.r.t. $Q$ over $\F$ if the following holds:
    \begin{equation}\label{eq:ipb-opt}
        g(\alpha) \geq \sup_{(q, p) \,\in\, \C_\F(P\,\|\,Q)} \big\{\, p ~\mid~ q \leq \alpha \,\big\}.
    \end{equation}
    We denote by $\B_{\F}(P ~\|~ Q)$ the set of all such IPB functions. 
    The function $g$ attaining equality for all $\alpha \in \overline{\Y}$ in \cref{eq:ipb-opt} is unique. We call it the \emph{tight IPB}, denoted as $g = \B^*_{\F}(P ~\|~ Q)$.
\end{definition}
An IPB as defined above describes only the upper boundary of the achievable region $\C_\F(P ~\|~ Q)$. We can recover a lower boundary as a generalized inverse $\smash{\check{g}^{-1}}(\alpha) \define \inf\{ \beta \mid \check{g}(\beta) \geq \alpha \}$ of an IPB $\check{g} \in \B_\F(Q ~\|~ P)$ for the reversed pair. We call $g$ and $\check{g}$ \emph{adjoint} to each other, and refer to a pair $(g, \check{g})$ as \emph{adjoint IPBs}. For notational simplicity, in the rest of the paper we mostly focus on upper bounds.

IPBs sharpen the bound in \cref{eq:envelope-bound} by leveraging the non-linear structure:
\begin{proposition}[Main property]\label{stmt:ipb-bound}
    If $g \in \B_{\F}(P ~\|~ Q)$ is an IPB, then for any $f \in \F$, we have:
    \begin{equation}\label{eq:ipb-bound}
        \E_P[f] \leq g(\E_Q[f]).
    \end{equation}
\end{proposition}
Thus, if we can solve the optimization problem in \cref{eq:ipb-opt}, we can sharpen any bound of the ubiquitous \cref{eq:envelope-bound} form. We provide illustrations of how IPBs improve over IPMs in \cref{fig:ipb}, and in \cref{sec:toy-example}. In this work, we make the following contributions:
\begin{enumerate}
    \item We introduce the concept of IPBs as a non-linear generalization of an IPM, and present its properties and relationship to the standard information-theoretic concepts such as the Neyman--Pearson test statistic and the hockeystick divergence.
    \item We provide algorithms for estimating IPBs with finite-sample guarantees, for two realistic settings: over all bounded functions, and over a learnable family of functions.
    \item We show how IPB-based analyses enable researchers and practitioners to obtain sharp bounds where such sharpness matters. We provide a substantially tighter bound on performance deterioration in inference-time alignment compared to an approach based on the KL divergence~\cite[see, e.g.,][]{beirami2025theoretical}, estimate performance without judges when changing a language model's inference engine~\cite{ritter2026llms}, and estimate proportional representation gaps in synthetic data, with, e.g., 15 p.p.\@ more precise estimates than a previous method from \citet{oesterling2024multi} in some cases. We also show additional applications to deriving tight bounds on generalization and privacy risk in \cref{app:extra-results}.
\end{enumerate}

\section{Properties of IPBs}
\label{sec:properties}

\paragraph{Geometry of boundaries.} First, we make several observations on the basic geometric properties of IPBs. The tight IPB $g = \B^*_\F(P ~\|~ Q)$ is always non-decreasing by construction for any function class~$\F$.
Moreover, if $\F$ is a convex set, then the tight IPB $g$ is concave as an upper boundary of a convex set. As a concave function on an interval is continuous on its interior, the tight IPB is also continuous on the interior of its domain.

For function families that are closed under complement, we have a simple closed form of the lower boundary of the achievable region $\C_\F(P ~\|~ Q)$. 
\begin{proposition}\label{stmt:flipped-ipb}
    Consider the set of all bounded functions:
    \vspace{-.1em}
    $$\F_1 \define \{ f \mid f: \X \to [0, 1] \}.$$
    Suppose that $\F \subseteq \F_1$ is closed under complement, i.e., $f \in \F$ implies $1 - f \in \F$, and let $g = \B^*_\F(P ~\|~ Q)$ and $\check{g} = \B^*_\F(Q ~\|~ P)$ be a pair of tight adjoint IPBs. Then, for all $\alpha \in [0, 1]$, the lower boundary of $\C_\F(P ~\|~ Q)$ has the following form:
    $
        \check{g}^{-1}(\alpha) = 1 - g(1 - \alpha).
    $
\end{proposition}
\paragraph{IPBs recover IPMs.} Given a pair of tight adjoint IPBs, which is equivalent to having the upper and lower boundaries of $\C_\F(P ~\|~ Q)$, we can recover an IPM:
\begin{proposition}[IPB to IPM]\label{stmt:ipb-to-ipm}
    Suppose that $g = \B^*_{\F}(P ~\|~ Q)$ and $\check{g} = \B^*_{\F}(Q ~\|~P)$ is a pair of tight adjoint IPBs, where $f \in \F$ maps $\X \to \Y \subseteq \R$. Then,
    we have:
    \begin{equation}
        D_\F(P, Q) = \max \left\{ \sup [ g(\alpha) - \alpha ], \, \sup [ \check{g}(\alpha) - \alpha ] \right\},
    \end{equation}
    with both suprema over $\alpha \in \bar \Y$.
    When either IPB is not tight, the equality becomes an upper bound.
\end{proposition}

\paragraph{IPBs collapse to IPMs for shift-invariant function classes.} Although in general IPBs contain more information than IPMs, for some function families the bound they imply degenerates:
\begin{restatable}{proposition}{ipbshiftinvariant}\label{stmt:ipb-shift-invariant}
    Suppose that $\F$ is shift-invariant and closed under negation, i.e., if $f \in \F$ then $f + c \in \F$ and $-f \in \F$ for any $c \in \R$, with all functions $f \in \F$ mapping $\X \to \R$. Suppose that $g \in \B_{\F}(P ~\|~ Q)$. Then, $g(\alpha) \geq \alpha + D_\F(P, Q)$.
\end{restatable}
We defer the proof of this and all other non-trivial statements in the main body to \cref{app:proofs}.
As an immediate implication, for a salient case of $\F$ being the set of Lipschitz functions,
IPBs are \emph{not} more useful than IPMs. At the same time, for other important classes of functions, such as sets of bounded functions or balls in reproducing kernel Hilbert spaces~\cite{gretton2012kernel}, IPBs both carry useful information and enable us to derive substantially sharper bounds, as we show next. For the rest of this section, we focus on the properties of IPBs for the family of all bounded functions $\F_1$ due to the fundamental connections of IPBs in this case to standard information-theoretic notions.

\paragraph{IPBs and the Neyman--Pearson lemma.} Over the set of bounded functions $\F_1$, the tight IPB is the ROC curve of the likelihood-ratio test \cite[see, e.g.,][Chap. 14]{polyanskiy2025information}:
\begin{proposition}\label{stmt:ipb-np}
    Let $P$ and $Q$ be probability distributions over $\X$ with densities $p$ and $q$ w.r.t.\ a common measure, e.g., $P + Q$, and let $L(x) \define \nicefrac{p(x)}{q(x)}$ be the likelihood ratio, with $L(x) = \infty$ if $q(x) = 0$. For a threshold $t \in [0, \infty]$ and $\eta \in [0, 1]$, consider the randomized likelihood-ratio test $\phi_{t, \eta}(x) \define \indic[L(x) > t] + \eta \, \indic[L(x) = t]$, with size and power:
    \begin{equation}
        \alpha(t, \eta) \define \E_Q[\phi_{t, \eta}] 
        , \qquad
        \beta(t, \eta) \define \E_P[\phi_{t, \eta}],
    \end{equation}
    respectively. Then, the tight IPB $g = \B^*_{\F_1}(P ~\|~ Q)$ is the ROC curve of the likelihood-ratio test: for every $\alpha \in [0, 1]$, there exist $t$ and $\eta$ with $\alpha(t, \eta) = \alpha$ such that the supremum in \cref{eq:ipb-opt} is attained by $\smash{\phi_{t, \eta}}$, thus $g(\alpha) = \beta(t, \eta)$.
\end{proposition}

\paragraph{IPBs and hockeystick divergence.}
IPBs are also closely related to the hockeystick divergence:
\begin{definition}
Let $\gamma \ge 0$.\footnote{When the hockeystick divergence is defined as an $f$-divergence, we need to restrict $\gamma \geq 1$. We do not use its $f$-divergence form, thus $\gamma \geq 0$ is valid within the scope of this work.} The $\gamma$-hockeystick divergence $\mathsf{E}_\gamma(P \| Q)$ is defined as:
$$\mathsf{E}_\gamma(P \| Q) \define \sup_{E \subseteq \X} \left[ P(E) - \gamma Q(E) \right] = \sup_{f \in \F_1} \left[ \E_P[f] - \gamma \E_Q[f] \right],$$
where the first supremum is over all measurable events $E$. 
\end{definition}
Specifically, we can obtain an IPB from the hockeystick divergence, and vice versa:
\begin{restatable}[IPB to Hockeystick]{proposition}{hockeystick}
\label{stmt:ipb-hockeystick}
Suppose that $g = \B^*_{\F_1}(P ~\|~ Q)$. Then, for any $\gamma \ge 0$:
\vspace{-.2em}
\begin{equation}\label{eq:ipb-to-hockeystick}
    \mathsf{E}_\gamma(P \| Q) = \sup_{\alpha \in [0, 1]} \left[ g(\alpha) - \gamma \alpha \right],
\end{equation}
which becomes an upper bound when $g$ is not tight. The following is a tight IPB, $g = \B^*_{\F_1}(P ~\|~ Q)$:
\vspace{-.2em}
\begin{equation}
    g(\alpha) = \inf_{\gamma \ge 0} \left[ \mathsf{E}_\gamma(P \| Q) + \gamma \alpha \right].
\end{equation}
\end{restatable}
In \cref{app:extra-results}, we also show how one could convert between IPBs and $f$-divergences.

\section{Toy Example: Model Performance under Subpopulation Shift}
\label{sec:toy-example}
\begin{wrapfigure}[20]{r}{0.35\linewidth}
\vspace{-1.4em}
\centering
\resizebox{\linewidth}{!}{%
\begin{tikzpicture}
\begin{axis}[
    xlabel={\Large $\E_Q[f]$},
    ylabel={\Large $\E_P[f]$},
    xmin=0, xmax=1, ymin=0, ymax=1,
    width=8cm, height=8cm,
    axis lines=box,
    axis equal image,
    axis line style={thick},
    enlargelimits=false,
    grid=major,
    grid style={dashed, gray!30},
    xtick={0, 0.2, 0.4, 0.6, 0.8, 1},
    ytick={0, 0.2, 0.4, 0.6, 0.8, 1},
    clip=true,
]
\addplot[ROCColor, ultra thick, name path=ipbupper]
    coordinates {(0,0) (0.10,0.40) (0.30,0.60) (1,1)};
\addplot[TVColor, ultra thick, name path=tvupper]
    coordinates {(0,0.30) (0.70,1) (1,1)};
\addplot[ROCColor, ultra thick, name path=ipblower]
    coordinates {(0,0) (0.70,0.40) (0.90,0.60) (1,1)};
\addplot[TVColor, ultra thick, name path=tvlower]
    coordinates {(0,0) (0.30,0) (1,0.70)};
\addplot[ChanceColor, thick, dashed] coordinates {(0,0) (1,1)};
\addplot[black, thin, densely dotted] coordinates {(0.5,0) (0.5,1)};
\tikzfillbetween[of=tvupper and tvlower]{TVColor, opacity=0.1};
\tikzfillbetween[of=ipbupper and ipblower]{ROCColor, opacity=0.1};
\end{axis}
\end{tikzpicture}%
}
\vspace{-1.6em}
\caption{IPB-based bounds (\textcolor{ROCColor}{\bf –}) and IPM-based bounds (\textcolor{TVColor}{\bf –}) for three groups with $(q_1,q_2,q_3)=(0.7,0.2,0.1)$ and $(p_1,p_2,p_3)=(0.4,0.2,0.4)$.
}
\label{fig:compact-example}
\end{wrapfigure}
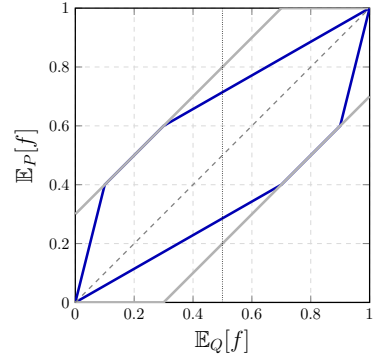
To illustrate the mechanics of IPBs, let us provide a simple example. Suppose we have a classifier that was validated on a population $Q$ but deployed on a population $P$ where the proportions of $k$ groups $G_1,\ldots,G_k$ differ, with $p_i\define P(G_i)$ and $q_i\define Q(G_i)$, whereas the performance $a_i\in[0,1]$ of the model within each group stays the same. The deployment performance is then $\E_P[f]=\sum_i a_ip_i$, with $f(a) = \sum_i a_i\indic[G_i]$, and we want to bound it given the validation performance $\alpha=\E_Q[f]=\sum_i a_iq_i$. Over the family of such functions:
$$\F = \Big\{ a \mapsto \sum_i a_i\indic[G_i] ~\mid~ a_i \in [0, 1] \Big\},$$
the IPM simplifies to the total-variation (TV) distance: $$D_{\F}(P,Q) = \sup_{a_i \in [0, 1]} \sum_i a_i (p_i-q_i) = \frac12\sum_i|p_i-q_i|.$$
By \cref{def:ipb}, the tight IPB $g = \B_{\F}^*(P \parallel Q)$ in this case is given by a solution of the linear program:
$$g(\alpha) \,=\, \max \sum_{i} a_i p_i \quad \text{ s.t. } \sum_i a_i q_i \leq \alpha,\ a_i \in [0, 1],$$
which is an instance of the classical \textit{fractional knapsack problem}. It has a known optimal solution~\cite[see, e.g.,][Chap. 3]{martello1987algorithms} that corresponds to greedily ``packing'' the groups until the ``knapsack'' is full, in the order of decreasing $r_i\define p_i/q_i$. Sorting $r_1\ge\cdots\ge r_k$, this means that we set $a_i=1$ for the first $m-1$ groups, and choose $a_m$ so that $\E_Q[f]=\alpha$, resulting in:
$$
    g(\alpha)=\textstyle\sum_{i<m}p_i+r_m\big(\alpha-\sum_{i<m}q_i\big), \quad \text{ for } \alpha\in \big[\sum_{i<m}q_i,\sum_{i\le m}q_i \big].
$$
We can obtain the lower boundary $\check g^{-1}$ using \cref{stmt:flipped-ipb}. \Cref{fig:compact-example} shows an example with three groups. For a validation performance of $0.5$ (dotted line), the IPM-based bounds give a deployment performance in $[0.2,0.8]$, whereas the IPB-based bounds give $[0.29,0.71]$. For $\alpha\ge0.7$, the IPM-based upper bound is vacuous, whereas $g(\alpha)<1$.

\section{Algorithms for IPBs}
\label{sec:algorithms}

We introduce new algorithms for two salient settings that we make use of in our applications.

\paragraph{Estimating the tight IPB over $\F_1$ with known $P$ and $Q$.}
First, we present \cref{alg:ipbnp} for computing an IPB over the set of all bounded functions when we know both $P$ and $Q$, and can sample from $Q$, e.g., when we want to compute an IPB between the distributions of generative models. The algorithm samples from $Q$ alone, and then uses importance weighting to recover expectations under $P$. Following \cref{stmt:ipb-np}, \cref{alg:ipbnp} sweeps thresholds $t$ over the observed likelihood ratios and returns a set of $\{(\hat{\alpha}_k, \hat{g}_k)\}_{k=0}^n$ points from the IPB. The empirical means of the likelihood ratios, however, can induce invalid boundaries due to the statistical noise. To correct for this, we also perform a normalization step that exploits the fact that the tight IPB in this setting satisfies $g(1) = 1$.
The algorithm has finite-sample convergence guarantees:
\begin{restatable}{theorem}{ipbnpfinite}\label{stmt:ipbnp-finite}
Suppose that $P \ll Q$ and $\nicefrac{dP}{dQ} \leq B$ almost surely under $Q$ for some $B <  \infty$. Let $\hat{g}_n$ denote the piecewise-linear interpolation of $\{(\hat\alpha_k, \hat{g}_k)\}_{k=0}^n$ returned by \cref{alg:ipbnp}, and let $g = \B^*_{\F_1}(P ~\|~ Q)$ be the tight IPB. Then, for any $\delta \in (0, 1)$, with probability at least $1 - \delta$ over the draw of samples from $Q$, we have:
$$ \sup_{\alpha\in[0,1]}\left|\hat g_n(\alpha)-g(\alpha)\right| = \mathcal{O}\left(B\sqrt{\tfrac{\log(2/\delta)}{n}}\right).$$
\end{restatable}
The algorithm is surprisingly simple for its power: (1) it enables us to have bounds of the form in \cref{eq:ipb-bound} over the \emph{entire} set of bounded functions without ever performing the actual optimization in the definition of IPBs over this set, and (2) it only requires having samples from $Q$. We can avoid the optimization because of the Neyman--Pearson connection in \cref{stmt:ipb-np}.

\begin{algorithm}[t]
\caption{IPB estimation for the family of bounded functions with known $P$ and $Q$}\label{alg:ipbnp}
\KwIn{Distributions $P$ and $Q$ with $P \ll Q$; sample size $n$.}
\For{$i = 1, \ldots, n$}{
    Sample $o_i \sim Q$, set $L_i \gets \nicefrac{dP}{dQ}(o_i)$ \\
}
$\hat \alpha_0 \gets 0, \ \hat{g}_0 \gets 0$\\
Sort $\{L_i\}_{i=1}^n$ in decreasing order as $L_{(1)} \geq L_{(2)} \geq \cdots \geq L_{(n)}$ \\
\For{$k = 1, \ldots, n$}{
    $\hat{\alpha}_k \gets k / n$, \ $\hat{g}_k \gets \frac{1}{n} \sum_{j=1}^{k} L_{(j)}$ \\
}
Normalize $\hat{g}_k \gets \nicefrac{\hat{g}_k}{\hat{g}_n}$ for all $k = 1, \ldots, n$ \\
\Return $\{(\hat{\alpha}_k, \hat{g}_k)\}_{k=0}^{n}$
\end{algorithm}

\paragraph{Estimating an IPB for learnable $\F$.}
Next, we present a more general algorithm, applicable even when we do not have access to the probability distributions $P$ and $Q$ themselves, and only have samples from them. In this case, unlike the previous algorithm, we need to solve the optimization problem in \cref{eq:ipb-opt}. In practice, it is convenient to use the Lagrangian form to avoid the constrained optimization, i.e., solve $\sup_{f \in \F} \left[ \E_P[f] - \lambda \E_Q[f] \right].$
When doing so, however, we can no longer explicitly control $\E_Q[f] \leq \alpha$. We make use of the following relationship:
\begin{proposition}[Lagrangian dual]\label{stmt:lagrangian}
Let $\lambda \geq 0$ be a \emph{dual variable}. We define the \textit{Lagrangian} as 
$h_\F(\lambda) \;\define\; \sup_{f \in \F} \left[ \E_P[f] - \lambda\,\E_Q[f]\right].$
Then, for any $f \in \F$ and $\alpha \geq \E_Q[f]$, we have: $$\E_P[f] \;\leq\; \inf_{\lambda \geq 0} [ h_\F(\lambda) + \lambda\,\alpha].$$
\end{proposition}

\begin{algorithm}[t]
\caption{IPB estimation for a learnable family $\F$}\label{alg:lagrangian}
\KwIn{Samples $X_P \sim P^{n_P}$, $X_Q \sim Q^{n_Q}$; family $\F$; grid $\Lambda = \{\lambda_1, \lambda_2, \ldots, \lambda_k\}$}
\For{$\lambda \in \Lambda \cup \{1\}$}{
    \vspace{-.5em}
    \begin{equation}\label{eq:algo-opt-lagrangian}
        \hat{h}(\lambda) \gets \max_{f \in \F} \left[ \frac{1}{n_P}\sum_{x \in X_P} f(x) - \lambda \frac{1}{n_Q} \sum_{x \in X_Q} f(x) \right]
    \end{equation}\\[-.75em]
}
$\hat g(\alpha) \gets \displaystyle\min_{\lambda\in\Lambda \cup \{1\}} \left[\hat{h}(\lambda)+\lambda\,\alpha\right]$\\
\Return $\hat g$.
\end{algorithm}

We present a procedure for estimating an IPB using optimization with a Lagrangian objective in \cref{alg:lagrangian}. Essentially, the algorithm solves $k$ optimization instances for a grid of $\lambda \in \Lambda$ values, and derives the IPB using \cref{stmt:lagrangian}. We also enforce $\lambda = 1$ to be part of the grid to ensure that the resulting estimates are at least as tight as an IPM.

In a practical application, it might not be clear how to choose $\Lambda$ in a principled way. As we explain in \cref{app:extra-results}, increasing $\lambda$ values in the grid beyond a certain point does not carry benefits. Thus, one principled way is to use a grid on $[0, \lambda_{\max}]$, where $\lambda_{\max}$ is an estimate of such a $\lambda$ after which we do not observe benefits, which happens when $h_{\F}(\lambda)$ is small enough. In \cref{alg:lambda-search} in the Appendix, we provide a heuristic algorithm to find $\lambda_{\max}$ using doubling search. Moreover, in \cref{alg:lagrangian-lower} in the Appendix, we also present a procedure analogous to \cref{alg:lagrangian} to get the lower boundary of the achievable region without the need to explicitly invert the adjoint IPB $\check{g} \in \B_\F(Q ~\|~P)$.

The algorithm provides a finite-sample convergence guarantee when $\F$ is regular enough. Each $\lambda \in \Lambda$ yields a valid upper bound by \cref{stmt:lagrangian}, thus the output of \cref{alg:lagrangian} is a valid IPB up to statistical and optimization errors. The following quantifies how far it is from the tight IPB:
\begin{restatable}{theorem}{lagrangianfinite}\label{stmt:lagrangian-finite}
Suppose $0 \in \F \subseteq \F_1$ is a convex family of bounded functions with their Rademacher complexity terms denoted as $\smash{\mathfrak{R}_{n_P}(P, \F)}$ and $\smash{\mathfrak{R}_{n_Q}(Q, \F)}$, that $P \ll Q$ with $\nicefrac{dP}{dQ} \leq B$ almost surely under $Q$, and that we can solve \cref{eq:algo-opt-lagrangian} up to at most $\varepsilon_{\mathrm{opt}} \geq 0$ error. Let $\Lambda \ni 1$ be a grid with $\lambda_{\max} \define \max \Lambda \geq B$ and covering radius $\eta_\Lambda \define \smash{\sup_{\lambda \in [0, \lambda_{\max}]}} \min_{\lambda' \in \Lambda} |\lambda - \lambda'|$. Then, for $\delta \in (0, 1)$, w.p. at least $1 - \delta$ over sampling of $X_P$ and $X_Q$, the output $\hat{g}$ of \cref{alg:lagrangian} satisfies:
$$
    \sup_{\alpha \in [0,1]}\left|\hat g(\alpha) - g(\alpha)\right| \;\leq\; \varepsilon_P + \lambda_{\max}\,\varepsilon_Q + \eta_\Lambda + \varepsilon_{\mathrm{opt}},
$$
where $g = \B^*_\F(P ~\|~ Q)$ is the tight IPB, and $\varepsilon_S \define 2\mathfrak{R}_{n_S}(S, \F) + \sqrt{\frac{\log(4/\delta)}{2n_S}}$ for $S \in \{P, Q\}$.
\end{restatable}
The bound combines four sources of error. The first two terms are statistical, and the latter two are structural due to the discretization of the dual variable, and optimization error, respectively. The discretization error quantifies the trade-off between tightness and the number of optimization oracle calls: e.g., for a uniform grid of $k$ points, $\eta_\Lambda = \nicefrac{\lambda_{\max}}{2(k-1)}$, so a discretization error of at most $\varepsilon$ requires $\mathcal{O}(\nicefrac{\lambda_{\max}}{\varepsilon})$ oracle calls. The last term is the optimization error, whose direction matters for validity: Any over-estimate of the maximum in \cref{eq:algo-opt-lagrangian} still yields a valid, even if loose, upper bound by \cref{stmt:lagrangian}, whereas an under-estimate can invalidate it.

In \cref{app:proofs}, we extend this analysis. First, we show that the dependence on $\lambda_{\max}$ is conservative---for any $\alpha$, we can replace it by the active dual variable at $\alpha$---thus deriving a tighter version of the analysis. Second, we show that for a general non-convex $\F$, the algorithm estimates a valid IPB, but it is separated from the tight one by an additional duality gap. Third, we extend the analysis to unbounded function families. Moreover, in \cref{app:extra-results}, we show how to modify the algorithm to provide a finite-sample certificate that is valid with high probability for simple function families. In \cref{app:exp-details}, we illustrate this certificate for the family of bounded linear functions, for which the optimization oracle is exact and tight Rademacher complexity expressions are available.

\section{Applications}
\label{sec:applications}

We present several applications of IPBs, chosen to showcase them in different settings. First, we show how one could leverage an IPB view to derive a tighter bound on a practically relevant quantity in \textit{alignment}.
Second, we show an empirical application to \textit{estimating language model capabilities under a change of inference backend}
using the connection of IPBs to the Neyman--Pearson lemma and \cref{alg:ipbnp}. Third, we show another empirical application to \textit{fairness in generative modeling},
this time using a learnable function family and \cref{alg:lagrangian}. 
Additionally, in \cref{app:extra-results}, we show theoretical applications to derive tight \textit{generalization} bounds and \textit{privacy risk} bounds in differentially private learning. We summarize the considered applications in \cref{tab:applications} in the Appendix.

\subsection{Bounding Capability Deterioration and Limits for Improvement in Alignment}
\label{sec:policy-comparison}

Alignment techniques~\cite[see, e.g.,][]{christiano2017deep,stiennon2020learning} aim to steer generative models using reward signals to improve their adherence to given requirements or human preferences. In alignment, one needs to balance maximizing reward and preserving existing capabilities.
An IPB between the original model and an aligned model can tell us the maximum increase or decrease of \emph{any} metric measuring an existing level of capabilities.
In this case study, we focus on best-of-$N$ (BoN) sampling~\citep{brown2024large,snell2024scaling}, a standard baseline for model alignment at inference time. We derive the closed-form IPB for BoN sampling and compare it against the bounds on capability changes from KL divergence, which is the standard analytical tool in this context~\cite{gao2023scaling,beirami2025theoretical,paes2026theoretical}.

Given a reference distribution $Q$, e.g., a language model, over an output space $\Y$, e.g., over $m$-token strings, and a scalar reward function $r: \Y \to \R$, the BoN procedure draws $N$ i.i.d.\ samples $y_1, \ldots, y_N \sim Q$ and returns the one with the highest reward, breaking any ties uniformly at random. Let $P_N$ denote the resulting distribution over outputs. We are interested in quantifying the difference in arbitrary bounded metrics $f \in \F_1$, e.g., error rate, task accuracy, etc., between the original $\E_Q[f]$ and the post-alignment $\E_{P_N}[f]$. To do so, we derive the tight IPB between $P_N$ and $Q$, which provides the following bound:
\begin{restatable}{theorem}{bonipb}\label{stmt:bon-ipb}
    For any bounded performance measure $f \in \F_1$, any reward function $r: \Y \to \R$ and any reference distribution $Q$:
    \vspace{-.2em}
    \begin{equation}
        \E_{y \sim P_N}[f(y)] \leq 1 - \left(1 - \E_{y \sim Q}[f(y)] \right)^N.
    \end{equation}
    The inequality is tight if $r$ guarantees no ties.
\end{restatable}
To illustrate this theoretical application of IPBs, we show a partial proof of this statement under the assumption that $r$ produces no ties. In \cref{app:proofs}, we show the remainder.
\begin{proof}
    Let $F_Q(v) \define \Pr_{y \sim Q}[r(y) \leq v]$ denote the cumulative distribution function of the reward under $Q$. As there are no ties, BoN outputs $y$ exactly when one of the $N$ draws equals $y$ and the other $N-1$ draws have reward below $r(y)$, so $dP_N(y) = N\,F_Q(r(y))^{N-1}\,dQ(y)$. 
    
    Thus, the likelihood ratio of $P_N$ over $Q$ is
    $L(y) \define \nicefrac{dP_N}{dQ}(y) = N\, F_Q(r(y))^{N-1}.$
    Moreover, the no-ties assumption implies that $r(y)$ has a continuous distribution under $Q$, thus $U \define F_Q(r(y))$ is uniform on $[0, 1]$ for $y \sim Q$, and, as a result, $L(y) = NU^{N-1}$. A useful fact that follows immediately is that for a threshold $t \in (0, N)$, we have $L(y) > t$ whenever $U > u_t$, for $u_t \define (t/N)^{\frac{1}{N-1}}$.
    
    By \cref{stmt:ipb-np}, the tight IPB is the ROC curve of the test $\indic[L(y) > t]$ (no randomization is needed, as $L(y)$ has no atoms), whose size under $Q$ and power under $P_N$ are:
    \begin{align*}
        \alpha(t) &= \Pr_{y \sim Q}[L(y) > t] = \Pr[U > u_t] = 1 - u_t, \\
        \beta(t) &= \Pr_{y \sim P_N}[L(y) > t] = \E_{y \sim Q}\left[L(y)\cdot\indic[L(y) > t]\right] \qquad \text{(change of measure $dP_N = L\,dQ$)} \\
        &= \E[NU^{N-1} \cdot \indic[U > u_t]]
        = \int_{u_t}^{1} N u^{N-1}\, du = 1 - u_t^N.
    \end{align*}
    Setting $u_t = 1 - \alpha(t)$ yields $g(\alpha) = 1 - (1-\alpha)^N$, and the statement follows by \cref{stmt:ipb-bound}.
\end{proof}

\begin{figure}[t]
\centering
\resizebox{0.75\linewidth}{!}{
\begin{minipage}[b]{0.45\linewidth}
\centering
\begin{tikzpicture}
\begin{axis}[
    xlabel={$\phantom{\E_P}N$},
    ylabel={$\E_{P_N}[f] - \E_Q[f]$},
    xmin=1, xmax=25,
    ymin=0, ymax=1.05,
    width=\linewidth, height=5.5cm,
    samples=120,
    axis lines=box,
    axis line style={thick},
    grid=major,
    grid style={dashed, gray!30},
    legend style={font=\tiny, draw=lightgray, fill=white, fill opacity=0.9, text opacity=1, at={(0.98,0.02)}, anchor=south east, row sep=-1pt},
    legend cell align={left},
]
\addplot[rocketA5, thick, domain=1:50] {1 - pow(1 - 0.05, x) - 0.05};
\addlegendentry{IPB, $\alpha=0.05$}
\addplot[rocketB5, thick, domain=1:50] {1 - pow(1 - 0.25, x) - 0.25};
\addlegendentry{IPB, $\alpha=0.25$}
\addplot[rocketC5, thick, domain=1:50] {1 - pow(0.5, x) - 0.5};
\addlegendentry{IPB, $\alpha=0.50$}
\addplot[rocketD5, thick, domain=1:50] {1 - pow(1 - 0.75, x) - 0.75};
\addlegendentry{IPB, $\alpha=0.75$}
\addplot[rocketE5, thick, domain=1:50] {1 - pow(1 - 0.95, x) - 0.95};
\addlegendentry{IPB, $\alpha=0.95$}
\addplot[sbgray, dashed, thick, domain=1:50, samples=200]
    {min(1, sqrt((ln(x) - (x-1)/x)/2))};
\addlegendentry{KL+Pinsker}
\addplot[black, dashed, thick, domain=1:50, samples=200]
    {min(1, sqrt(1 - exp(-(ln(x) - (x-1)/x))))};
\addlegendentry{KL+BH}
\end{axis}
\end{tikzpicture}
\end{minipage}%
\hspace{2em}
\begin{minipage}[b]{0.45\linewidth}
\vspace{2em}
\centering
\begin{tikzpicture}
\begin{axis}[
    xlabel={$\E_Q[f]$},
    ylabel={$\E_{P_N}[f]$},
    xmin=0, xmax=1,
    ymin=0, ymax=1,
    width=\linewidth, height=5.5cm,
    samples=200,
    axis lines=box,
    axis line style={thick},
    grid=major,
    grid style={dashed, gray!30},
    legend style={font=\tiny, draw=lightgray, fill=white, fill opacity=0.9, text opacity=1, at={(0.98,0.02)}, anchor=south east, row sep=-1pt},
    legend cell align={left},
]
\addplot[sbgray, thick, dashed, domain=0:1, forget plot] {x};
\addlegendentry{$N=2$}
\addplot[rocketA4, thick, domain=0:1] {1 - pow(1-x, 2)};
\addlegendentry{$N=3$}
\addplot[rocketB4, thick, domain=0:1] {1 - pow(1-x, 3)};
\addlegendentry{$N=5$}
\addplot[rocketC4, thick, domain=0:1] {1 - pow(1-x, 5)};
\addlegendentry{$N=10$}
\addplot[rocketD4, thick, domain=0:1] {1 - pow(1-x, 10)};
\addlegendentry{$N=15$}
\addplot[rocketE5, thick, domain=0:1] {1 - pow(1-x, 15)};

\addplot[rocketA4, thick, domain=0:1] {pow(x, 2)};
\addplot[rocketB4, thick, domain=0:1] {pow(x, 3)};
\addplot[rocketC4, thick, domain=0:1] {pow(x, 5)};
\addplot[rocketD4, thick, domain=0:1] {pow(x, 10)};
\addplot[rocketE5, thick, domain=0:1] {pow(x, 15)};

\end{axis}
\end{tikzpicture}
\end{minipage}
}
\vspace{-.5em}
\caption{\textbf{IPBs provide tighter bounds on performance differences for best-of-$N$ sampling.} \textit{Left:} Upper bound on the gap $\E_{P_N}[f] - \E_Q[f]$, for $\E_Q[f] \leq \alpha$, as a function of $N$. The IPB-based bounds from \cref{stmt:bon-ipb} are in solid lines. We show KL-based bounds via Pinsker's inequality and the Bretagnolle--Huber (BH) inequality in dashed lines. \textit{Right:} IPBs for several fixed values of $N$.}
\label{fig:bon}
\vspace{-.5em}
\end{figure}
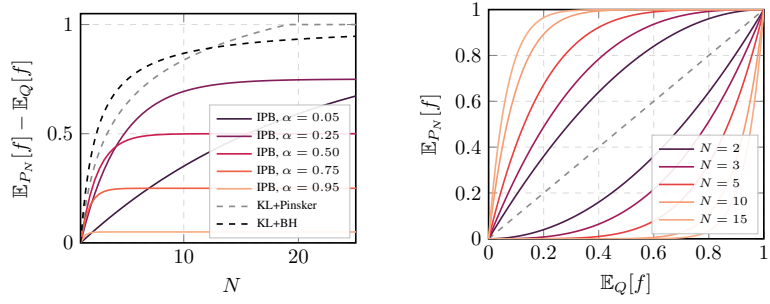

\paragraph{Comparison with KL-based bounds.} Prior analyses of BoN often use $D_\mathsf{KL}(P_N \| Q)$~\citep[see, e.g.,][]{mroueh2024information,beirami2025theoretical}. A standard approach is to bound it as $D_\mathsf{KL}(P_N \| Q) \leq \log N - (N-1)/N$.
We can then plug it into, e.g.,
Pinsker's inequality:
\vspace{-.2em}
\begin{equation}
    \E_{y \sim P_N}[f(y)] \leq \E_{y \sim Q}[f(y)] + \sqrt{\tfrac{1}{2}D_\mathsf{KL}(P_N \| Q)},
\end{equation}
for any $f \in \F_1$. Clearly, \cref{stmt:bon-ipb} yields a bound that is tighter due to its dependence on $\E_Q[f]$. We show the illustration in \cref{fig:bon}. As an example, at $N=10$, the IPB-based bound is 14 p.p. tighter for the worst-case $\E_Q[f]$, and 83 p.p. tighter at the extremes where $\E_Q[f] \approx 0$ or $\E_Q[f] \approx 1$.

\subsection{Judge-Free Estimation of Model Properties in Inference Engines}
\label{sec:prompt-audit}

Language models are often served through inference engines such as vLLM~\citep{kwon2025vllm}. Such deployments slightly alter their probability distribution compared to an original development backend~\citep{ritter2026llms}. If the original model has been audited, e.g., for its safety, on the development backend, the deployment version could deviate from it. IPBs can provide an estimate of closeness between the original model and its inference-time counterpart across all evaluation metrics from a chosen class, without requiring those metrics to be re-evaluated.

Let $M$ be a distribution of outputs from a language model under one backend, e.g., huggingface \texttt{transformers} \citep{wolf2020transformers}, and $M'$ be the same distribution under an inference backend, e.g., on vLLM. Let $D$ be a dataset of test inputs, $Q$ denote the joint distribution of $(x, y)$ obtained by drawing $x \sim \mathsf{Unif}[D]$ and $y \sim M(\cdot \mid x)$, and let $P$ denote the analogous distribution under $M'(\cdot \mid x)$. Any evaluation metric $f(x, y) \in [0, 1]$ has expectation $\E_Q[f]$ on the original backend, and $\E_P[f]$ under the inference backend. We assume that as part of an audit we have estimated the score $\E_Q[f]$ for the original model. A pair of adjoint IPBs $g \in \B_{\F_1}(P ~\|~ Q)$ and $\check{g} \in \B_{\F_1}(Q~\|~ P)$ then estimates $\smash{\check{g}^{-1}}(\E_Q[f]) \leq \E_P[f] \leq g(\E_Q[f])$ for any bounded evaluation metric $f$ without any additional scoring under $P$. Therefore, IPBs could enable deployers to avoid both drawing additional samples and scoring the outputs with expensive LLM-as-judges.

\paragraph{Experiment setup.}
We use IPBs to estimate the safety behavior of the \texttt{Llama-3.1-8B-Instruct} model~\citep{touvron2023llama} under three different backends. We use \cref{alg:ipbnp}, and take the original model $M$ to be the Llama model served via huggingface \texttt{transformers} with \texttt{bfloat16} precision, and rolled out over $m = 512$ tokens. This corresponds to the distribution $Q$. We consider three deployment variants $M'$ corresponding to distributions $P$: (i) a regular vLLM deployment, (ii) FP8 quantization~\citep{kuzmin2022fp8}, and (iii) AWQ-INT4 quantization~\citep{lin2024awq}. We evaluate all versions on three safety benchmarks: HarmBench~\citep{mazeika2024harmbench} (random subsample, $n = 200$), StrongREJECT~\citep{souly2024strongreject} (random subsample, $n = 200$), and JailbreakBench~\citep{chao2024jailbreakbench} ($n = 100$). In a slight deviation from the recommended evaluation protocols that prescribe specific judges and rubrics in different benchmarks, we use the rate of compliance with harmful requests (lower is better) as judged by Anthropic's \texttt{claude-haiku-4-5} model for simplicity and consistency. In a practical deployment, it would be sufficient to measure scores with a judge only for the original model. We compare the scores to the IPB bounds obtained with \cref{alg:ipbnp}, and, as a baseline, to bounds based on KL divergence and either Pinsker's or BH inequality (whichever is tighter), as in the previous section. See more details in \cref{app:exp-details}.

\begin{figure}[t]
\centering
\small
\includegraphics[width=\linewidth]{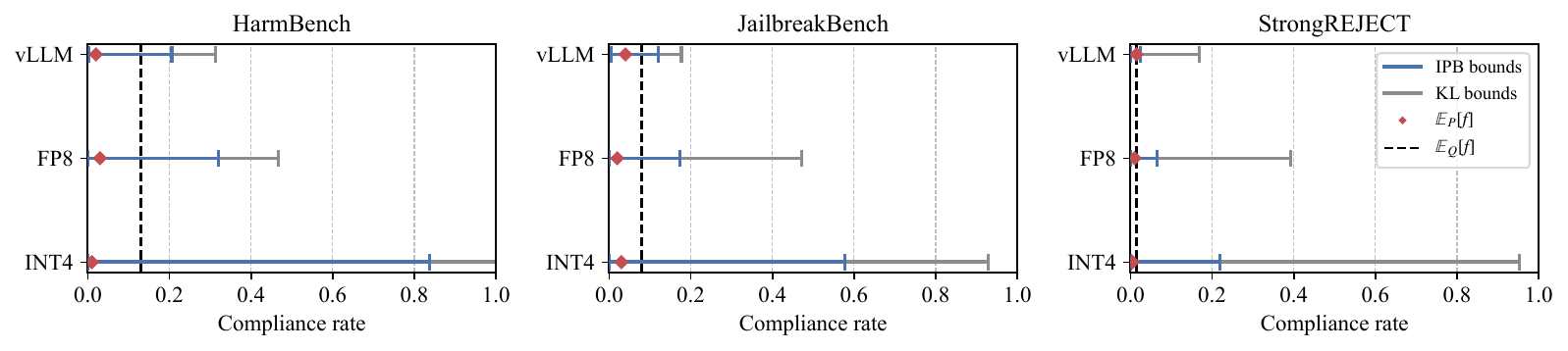}
\caption{\textbf{Unlike KL-based bounds, IPBs enable accurate judge-free estimation of benchmark scores when deploying a model to vLLM.} We show the compliance rate to harmful requests (lower is better) under different benchmarks for three different deployment backends, as well as the estimated bounds on the compliance rate from IPBs
between the original model $Q$ and its version under a given backend $P$, computed using \cref{alg:ipbnp}, and the bounds from the estimated KL divergence.}
\label{tab:benchmark}
\end{figure}

\paragraph{Experiment results.}
\cref{tab:benchmark} shows the outcomes. We show the full estimated IPB curves for all benchmarks and backends in \cref{fig:benchmark-curves} in \cref{app:extras}. When quantizing the model, the bounds from IPBs are loose. In some cases, however, e.g., INT4 on HarmBench, the lower bound is almost reached. This suggests that a wide interval from an IPB could indicate that a model should be re-audited. When only moving from the huggingface engine to vLLM, the IPB bounds remain fairly narrow, e.g., within 5 percentage points from the original value in the case of StrongREJECT. 
The results show that IPBs could enable judge-free performance estimation in certain settings.

\begin{figure}[t]
\centering
\includegraphics[width=0.72\linewidth]{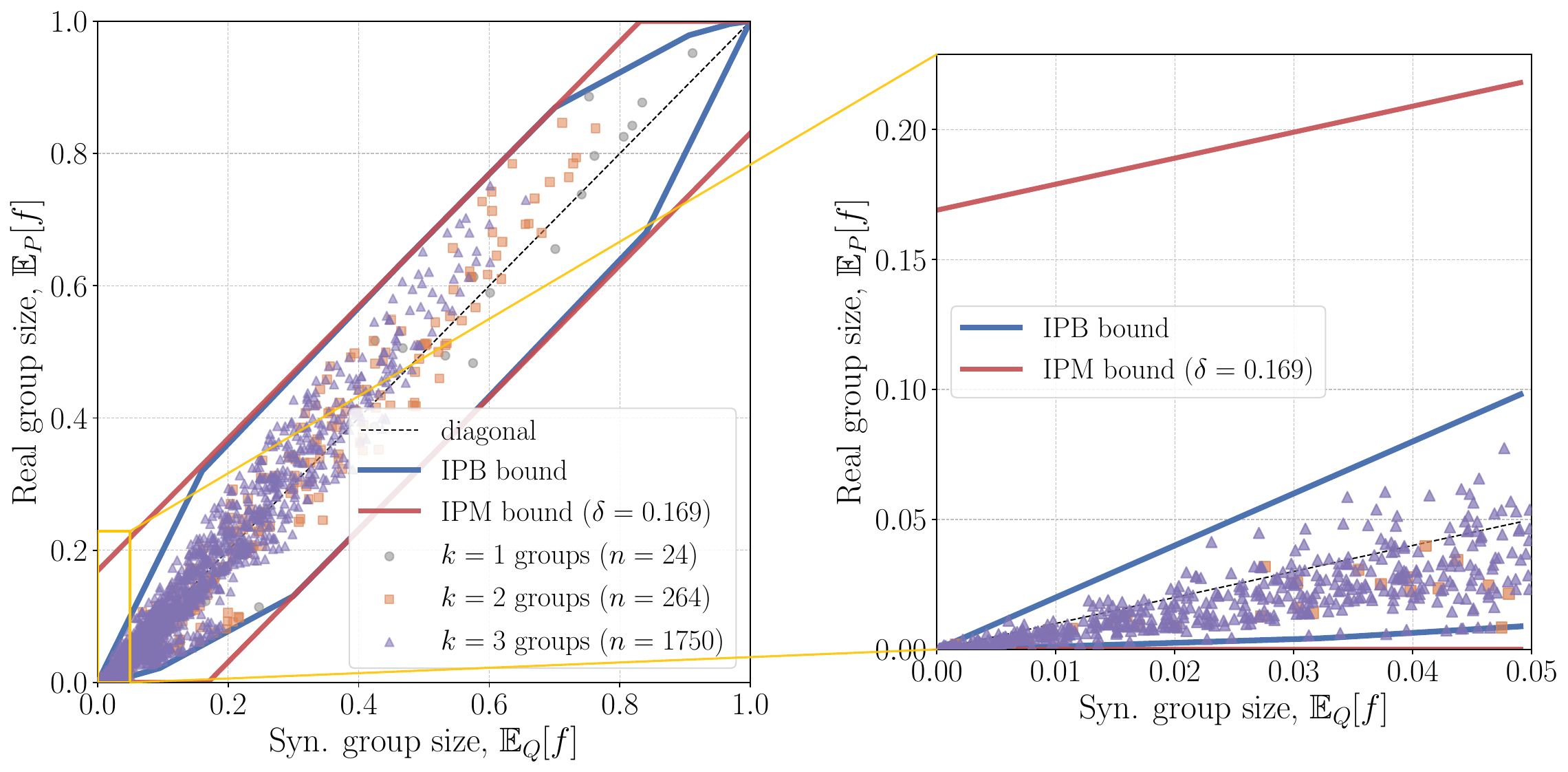}
\caption{\textbf{Estimated IPB bounds for multi-group proportional representation are substantially tighter for rare groups, whereas the bounds based on an IPM are often loose.} We show the bounds on the representation of intersectional groups defined by depth-3 decision trees, obtained from an IPM using the algorithm from \citet{oesterling2024multi}. We also show an upper bound and lower bound from IPBs 
computed using \cref{alg:lagrangian} and \cref{alg:lagrangian-lower}, respectively. Each point shows the prevalence of one such group in the synthetic dataset (x axis) and the real population (y axis). \textit{Left:} Over the entire $[0, 1]$. \textit{Right:} Zoom into the $[0, 0.05]$ region of rare groups.}
\label{fig:repr-cert}
\vspace{-.5em}
\end{figure}

\subsection{Estimating Multi-Group Proportional Representation Gaps for Rare Groups}
\label{sec:repr-cert}

Synthetic or distilled datasets and search retrieval results are increasingly used as proxies for large real datasets in various analytic workflows. Whenever we work with such \emph{surrogate data} that is a proxy for real data, one desirable property is \emph{proportional representation}~\citep{oesterling2024multi,jung2025multi}. Formally, denote by $P$ a reference distribution of records, e.g., a sample of real face images, and by $Q$ a distribution induced by the surrogate dataset, e.g., synthetic face images. For every demographic group defined by a group-membership indicator $f(x)$, we want to ensure that its real prevalence, $\E_P[f]$, closely matches its prevalence in the surrogate data, $\E_Q[f]$.

One metric that captures such differences is the IPM over a family of group-membership indicators $\F \subseteq \F_1$. Such an IPM is known as the \textit{multi-group proportional representation} (MPR) gap~\citep{oesterling2024multi}:
$\sup_{f \in \F}| \E_P[f] - \E_Q[f]|.$ The original use case of MPR has been as the objective in a set-valued optimization to construct a representative surrogate dataset. We consider a different goal: Given a surrogate dataset, an analyst wants to know whether the prevalence of a certain group they observe in the surrogate dataset is similar to the prevalence of the same group in the actual data. As we have seen before, bounds based on IPMs become vacuous when the prevalence is either low or high. In this setting, we can use IPBs to provide bounds on the representation of \emph{arbitrary} groups defined by a family of indicators $\F$ without the need to explicitly test the representation of each.
Specifically, say we have an adjoint pair of IPBs $g \in \B_\F(P ~\|~ Q)$ and $\check{g} \in \B_\F(Q ~\|~ P)$. Then, having observed a certain prevalence of a given group in the surrogate dataset $Q$, we can obtain bounds on its prevalence in the real data as $\check{g}^{-1}(\E_Q[f]) \leq \E_P[f] \leq g(\E_Q[f])$.

\paragraph{Experiment setup.} We demonstrate the applicability of IPBs to multi-group proportional representation for the use case of generating a synthetic dataset of a Census population. Let $\X = \{0,1\}^d$ be a binary demographic feature space. We define the class $\F_k$ to be all group-membership indicators obtained from decisions of some depth-$k$ decision tree over $\X$. This class contains every $k$-way intersectional group (e.g., ``\texttt{white}\,$\wedge$\,\texttt{married}\,$\wedge$\,\texttt{over 60 years old}'') as a special case.

We use California 2018 Census data obtained via the \texttt{folktables} package~\cite{ding2021retiring}. We binarize $d=12$ demographic features covering sex, race/ethnicity, country of origin, disability status, age brackets, marital status, income, and educational status. We take $P$ to be the held-out test split ($n_P = 75{,}763$) and $Q$ a synthetic dataset ($n_Q = n_P$). We generate the synthetic dataset using CTGAN~\cite{xu2019modeling} on the remaining real dataset $(n \approx 300{,}000)$. We use doubling search to find $\lambda_{\max}$, as described in \cref{app:extra-results}. Then, we run \cref{alg:lagrangian} with a grid $\Lambda$ of 10 values on $[0, \lambda_{\max}]$, including $\lambda = 1$ (see \cref{app:exp-details} for all hyperparameters). To practically compare the obtained bounds to ground truth for the purpose of the experiment, we set a relatively low $k = 3$, corresponding to all intersectional groups defined by depth-3 decision trees (see an example above), filtering out those that contain fewer than 10 records for stability. This results in 2042 groups. As a baseline, we estimate the MPR gap based on an IPM as $\delta = \sup_{f \in \F_k} |\E_P[f] - \E_Q[f]|,$ as done by \citet{oesterling2024multi}. In \cref{app:exp-details}, we also show how one could obtain a certified IPB for the family of bounded linear functions, also studied by \citet{oesterling2024multi}.

\paragraph{Experiment results.}
\cref{fig:repr-cert} shows the results. The bounds obtained with IPBs using our \cref{alg:lagrangian} are empirically tighter than those obtained from the corresponding IPM. Specifically, for rare groups, the lower bound from an IPM on the prevalence in the real dataset is always vacuous, unlike with IPBs. For the smallest group, the gap between an IPB-based bound and an IPM can be as high as $500\times$. The same results hold in replications with two other random train-test splits (see \cref{app:exp-details}). 
These results show that IPB curves can serve as a measure of fidelity for surrogate data. The analysts working with surrogate data could use IPBs to understand how well group-based queries performed on the surrogate dataset replicate in the real data.

\section{Related Work}
The objects related to IPBs have appeared previously across different contexts, but have not been unified as a generalization of IPMs as we develop in this work. In particular, the IPB over the family of bounded functions, i.e., the ROC of the Neyman--Pearson test, has been studied and related to the hockeystick divergence and $f$-divergences multiple times independently, e.g., by \citeauthor{polyanskiy2010channel} in his Ph.D. thesis, \citeyear{polyanskiy2010channel}; by \citet{reid2011information} in the context of risk in binary experiments; by \citet{sason2016f} in the form of \emph{relative information spectrum}; by \citet{agrawal2021optimal} in the context of optimal bounds between $f$-divergences and IPMs; and by \citet{dong2022gaussian} in the context of differential privacy.
We generalize beyond the family of all bounded functions, and focus on tools and algorithms for obtaining sharp bounds in applications where this sharpness matters.

A highly related problem is that of Neyman--Pearson (NP) classification~\cite[see, e.g.,][]{scott2005neyman,rigollet2011neyman,tong2018neyman}. The goal in NP classification is to learn a binary classifier with a given constraint on false-positive rate, which is equivalent to estimating a point on the tight IPB over a learnable family of binary classifiers. In contrast, our goal is to bound the expectations in broader function families.

\section{Concluding Remarks}
\label{sec:conclusions}
We introduced integral probability boundaries (IPBs), a functional generalization of integral probability metrics. We showed that IPBs are a useful framing and tool both for deriving tighter bounds in theoretical analyses, and for obtaining precise estimates at the extremes when comparing probability distributions in practical applications. Our hope is that this framework will help both researchers and practitioners to analyze and audit aspects of trustworthiness such as safety and fairness in machine learning, where performance at the extremes is important, in a more precise and inclusive way.

\paragraph{Provable guarantees.} 
IPBs provide only approximate bounds in some settings, limiting their applicability when we need provable guarantees, as the quality of the estimate in \cref{alg:lagrangian} depends on the optimality of the optimization oracle. 
Exploring settings where we can obtain provable guarantees, e.g., using convex relaxations, is a potentially useful direction for future work. 

\paragraph{Alternative computational approaches.} Exploring alternatives to \cref{alg:lagrangian} for appropriate function families is another promising direction for future work. For instance, we could estimate an IPB by solving the problem in \cref{eq:ipb-opt} directly via the NP classification and then deriving the NP-ROC curves~\cite{tong2018neyman}, or via projected gradient descent for differentiable families.

\paragraph{When IPMs suffice.} Finally, although IPBs are a drop-in replacement for IPMs, not every application requires the added complexity of IPBs. First, as shown in \cref{sec:properties}, IPBs do not provide improvements over IPMs for shift-invariant function families. For one salient case, this includes the family of Lipschitz functions; therefore, IPBs do not improve bounds based on Wasserstein distance.

Second, in some applications, it is more convenient to summarize the difference between distributions using a single number. Consider the goal of minimizing the difference between two distributions. 
As an example, \citet{oesterling2024multi} study the problem of constructing a representative surrogate dataset that minimizes the representation gaps compared to the reference distribution $P$ across all groups within a given family $\F$, i.e., modeling the surrogate dataset as a probability distribution $Q$, the goal is to solve $\min_Q D_{\F}(P, Q)$. In such a case, it is convenient to work with a scalar-valued discrepancy $D_{\F}(P, Q)$ as opposed to a set of points on the boundary as part of the optimization problem. Instead, IPBs are likely most useful when fine-grained differences at the extremes matter, such as in audits and analyses, as in the applications we study in \cref{sec:policy-comparison,sec:prompt-audit,sec:repr-cert}.

\section*{Acknowledgements} The computing resources were provided by Modal. SA was partially supported by the NSERC Discovery Grant. BK was partially supported by the Swiss National Science Foundation under award numbers 10003518 and 237378, and by the SYNTHIA project. SYNTHIA (Synthetic Data Generation framework for integrated validation of use cases and AI healthcare applications) project is supported by the Innovative Health Initiative Joint Undertaking (IHI JU) under grant agreement No. 101172872. Thus, the project is partially funded by the European Union, the private members, and those contributing partners of the IHI JU. Views and opinions expressed are however those of the authors only and do not necessarily reflect those of the aforementioned parties. Neither of the aforementioned parties can be held responsible for them. 

\bibliographystyle{unsrtnat}
\bibliography{main}

\clearpage

\appendix
\crefalias{section}{appendix}

\clearpage
\section{Omitted Proofs}
\label{app:proofs}

\subsection{Proofs for Section~\ref{sec:properties}}

We start with the proofs of the basic properties.

\begin{restatable}{proposition}{ipbconcave}\label{stmt:ipb-concave}
    Suppose that $\F$ is convex, i.e., $\lambda f_1 + (1-\lambda) f_2 \in \F$ whenever $f_1, f_2 \in \F$, for any $\lambda \in [0,1]$. Then, $g = \B^*_\F(P ~\|~ Q)$ is concave on $\overline{\Y}$, with the convention $\sup \emptyset = -\infty$.
\end{restatable}

\begin{proof}
Denote $\F_\alpha \define \{f \in \F \mid \E_Q[f] \leq \alpha\}$, so that $g(\alpha) = \sup_{f \in \F_\alpha} \E_P[f]$. Fix $\alpha_1, \alpha_2 \in \overline{\Y}$, $\lambda \in [0, 1]$, and let $\alpha_\lambda \define \lambda \alpha_1 + (1-\lambda)\alpha_2$. Observing that for any $f_1 \in \F_{\alpha_1}$ and $f_2 \in \F_{\alpha_2}$, the function $f_\lambda \define \lambda f_1 + (1-\lambda) f_2$ lies in $\F$ by convexity and satisfies $\E_Q[f_\lambda] \leq \alpha_\lambda$ by linearity, we have:
$$ g(\alpha_\lambda) \ge \E_P[f_\lambda] = \lambda \E_P[f_1] + (1-\lambda)\E_P[f_2]. $$
Taking the supremum over $f_1 \in \F_{\alpha_1}$ and $f_2 \in \F_{\alpha_2}$ on the r.h.s.\@ yields $g(\alpha_\lambda) \ge \lambda g(\alpha_1) + (1-\lambda) g(\alpha_2)$.
\end{proof}

\ipbshiftinvariant*

\begin{proof}
Fix $\alpha \in \R$. As before, we set $\F_\alpha = \{ f \in \F ~\mid~ \E_Q[f] \leq \alpha \}$. Observe that for any $f \in \F$, there also exists $h \in \F$ that is its shifted version: $h = \alpha + f - \E_Q[f]$. Regardless of $f$, the shifted version $h$ satisfies $\E_Q[h] = \alpha$, thus $h \in \F_\alpha$. Therefore, denoting by $\F'_\alpha = \{ \alpha + f - \E_Q[f] \mid f \in \F \}$, we have that $\F'_\alpha \subseteq \F_\alpha$.

As a result, $\sup_{h \in \F'_\alpha} \E_P[h] \leq \sup_{f \in \F_\alpha} \E_P[f] \leq g(\alpha)$, where the last inequality is the definition of an IPB. Further, we can compute the supremum over $\F'_\alpha$:
\[
    \sup_{h \in \F'_\alpha} \E_P[h] = \sup_{f \in \F}\big[ \alpha + \E_P[f] - \E_Q[f]\big] = \alpha + D_\F(P, Q),
\]
where the last equality uses closure under negation: $\sup_{f \in \F}(\E_P[f] - \E_Q[f]) = \sup_{f \in \F}|\E_P[f] - \E_Q[f]| = D_\F(P, Q)$, as for any $f \in \F$ the function $-f \in \F$ achieves $\E_P[-f] - \E_Q[-f] = \E_Q[f] - \E_P[f]$.
Thus, we have that $\alpha + D_\F(P, Q) \leq g(\alpha)$ for any $\alpha \in \R$.
\end{proof}

\begin{remark}
    In \cref{stmt:ipb-shift-invariant}, we can relax the conditions on the family by removing the assumption that it is closed under negation. In that case, even though the IPB does not reduce to the standard IPM, it reduces to an asymmetric version of an IPM:
    $$g(\alpha) \geq \alpha + \sup_{f \in \F} [ \E_P[f] - \E_Q[f] ].$$
    (Note the absence of the absolute value.)
\end{remark}

\hockeystick*

\begin{proof}
    For any test $f \in \F_1$, by the IPB property we have:
    $$\E_P[f] - \gamma \E_Q[f] \le g(\E_Q[f]) - \gamma \E_Q[f].$$
    As the tight IPB $g$ is concave and non-decreasing, we can maximize over $\alpha = \E_Q[f] \in [0, 1]$, which yields:
    $$g(\E_Q[f]) - \gamma \E_Q[f] \leq \sup_{\alpha \in [0, 1]} \left[ g(\alpha) - \gamma \alpha \right].$$
    Therefore, \cref{eq:ipb-to-hockeystick} follows from the definition of the hockeystick divergence.
    
    To show tightness when $g$ is a tight IPB, we use a sandwich bound. First, notice that for every $\alpha \in [0, 1]$:
    $$g(\alpha) - \gamma \alpha
    \;=\; \sup_{f \in \F_\alpha}\!\left[\E_P[f] - \gamma \alpha\right]
    \;\le\; \sup_{f \in \F_\alpha}\!\left[\E_P[f] - \gamma \E_Q[f]\right]
    \;\le\; \mathsf{E}_\gamma(P \| Q),$$
    where the first inequality uses $-\gamma \alpha \le -\gamma \E_Q[f]$, as $\gamma \ge 0$ and $\E_Q[f] \le \alpha$, and the second drops the constraint. Then, taking the supremum on the left gives the
  matching lower bound, so the inequality holds with equality.

    Next, we verify that $\tilde g(\alpha) \define \inf_{\gamma \ge 0}[ \mathsf{E}_\gamma(P\|Q) + \gamma \alpha ]$ defines an IPB. For any $f \in \F_\alpha$ and any $\gamma \ge 0$, the definition of the
   hockeystick divergence gives $\E_P[f] - \gamma \E_Q[f] \le \mathsf{E}_\gamma(P\|Q)$, so after rearranging we have:
    $$\E_P[f] \;\le\; \mathsf{E}_\gamma(P\|Q) + \gamma \E_Q[f] \;\le\; \mathsf{E}_\gamma(P\|Q) + \gamma \alpha.$$
    As this holds for every $\gamma \ge 0$, taking the infimum on the right yields $\E_P[f] \le \tilde g(\alpha)$, so $\tilde g \in \B_{\F_1}(P ~\|~ Q)$. As $g$ is the tight IPB, this implies $\tilde g \geq g$.

    Finally, we show the reverse inequality. For any $\alpha \in [0, 1]$, \cref{stmt:ipb-np} provides $t$ and $\eta$ such that the test $\phi_{t, \eta}$ satisfies $\alpha(t, \eta) = \alpha$ and $\beta(t, \eta) = g(\alpha)$.

    Because $\E_P[f] - t\E_Q[f] = \E_Q[(L - t) f]$, the supremum of this expression over $f \in \F_1$ is achieved by any test that is $1$ on $\{L > t\}$ and $0$ on $\{L < t\}$. As $\phi_{t, \eta}$ satisfies this condition, it is a maximizer in the definition of $\mathsf{E}_t(P \| Q)$, and therefore:
    $$ \mathsf{E}_t(P\|Q) \;=\; \beta(t, \eta) - t \, \alpha(t, \eta) \;=\; g(\alpha) - t\alpha. $$

    By the definition of $\tilde g(\alpha)$ as an infimum, we have $\tilde g(\alpha) \le \mathsf{E}_t(P\|Q) + t\alpha = g(\alpha)$, confirming that $\tilde g = g$.

  \end{proof}
  
The above result can also be seen as a convex conjugate of $g$:
\begin{remark}
    Suppose that $g \in \B_{\F_1}(P ~\|~ Q)$. Then, for any $\gamma \geq 0$,
    $$\mathsf{E}_\gamma(P \| Q) \leq 1 + (1-g)^{*}(-\gamma),$$
    where $(\cdot)^*$ denotes the convex conjugate operation, with equality if $g = \B^*_{\F_1}(P ~\|~ Q)$ is the tight IPB.
\end{remark}

\begin{proof}
Let $h(\alpha) = 1-g(\alpha)$. By the definition of the convex conjugate, with the supremum taken over the domain of $g$:
$$h^*(y) = \sup_{\alpha \in [0, 1]} (y\alpha - h(\alpha))$$
Evaluating at $y = -\gamma$:
$$(1-g)^{*}(-\gamma) = \sup_{\alpha \in [0, 1]} (-\gamma\alpha - (1-g(\alpha)))$$
$$= \sup_{\alpha \in [0, 1]} (g(\alpha) - \gamma\alpha - 1)$$
$$= \bigg( \sup_{\alpha \in [0, 1]} (g(\alpha) - \gamma\alpha) \bigg) - 1$$
Substituting this result back, we get:
$$1 + (1-g)^{*}(-\gamma) = 1 + \bigg( \sup_{\alpha \in [0, 1]} (g(\alpha) - \gamma\alpha) \bigg) - 1 = \sup_{\alpha \in [0, 1]} (g(\alpha) - \gamma\alpha).$$
The claim follows from \cref{eq:ipb-to-hockeystick}, which holds with equality for the tight IPB.
\end{proof}

\subsection{Analysis of Algorithm~\ref{alg:ipbnp}}

Before proceeding, we start with some technical lemmas.
\begin{lemma}[Normalization]\label{lem:normalization}
Let $g \colon [0,1] \to [0,1]$ satisfy $g(1) = 1$, and let $\tilde g \colon [0,1] \to [0, \infty)$ be non-decreasing with $\sup_{\alpha \in [0,1]} |\tilde g(\alpha) - g(\alpha)| \leq r$. Define $\hat g \define \nicefrac{\tilde g}{\tilde g(1)}$ if $\tilde g(1) > 0$, and $\hat g \define \tilde g$ otherwise. Then:
$$\sup_{\alpha \in [0,1]} |\hat g(\alpha) - g(\alpha)| \leq 2r.$$
\end{lemma}
\begin{proof}
Fix $\alpha \in [0,1]$. If $\tilde g(1) = 0$, then $\tilde g \equiv 0$ as $\tilde g$ is non-negative and non-decreasing, so $\hat g = \tilde g$ and the claim follows from the assumption on $\tilde g$. Otherwise, factoring $\nicefrac{\tilde g(\alpha)}{\tilde g(1)}$ out of both terms, we have:
$$\hat g(\alpha) - \tilde g(\alpha) = \frac{\tilde g(\alpha)}{\tilde g(1)} - \tilde g(\alpha) = \frac{\tilde g(\alpha)}{\tilde g(1)}\bigl(1 - \tilde g(1)\bigr) = \hat g(\alpha)\bigl(1 - \tilde g(1)\bigr),$$
where $\hat g(\alpha) \in [0,1]$ as $0 \leq \tilde g(\alpha) \leq \tilde g(1)$ by monotonicity. Using this fact, and by the triangle inequality, we have:
\begin{align*}
|\hat g(\alpha) - g(\alpha)|
&\leq |\hat g(\alpha) - \tilde g(\alpha)| + |\tilde g(\alpha) - g(\alpha)| \\
&\leq |1 - \tilde g(1)| + r \\
&= |g(1) - \tilde g(1)| + r \leq 2r,
\end{align*}
where the last line uses $g(1) = 1$ and the assumption on $\tilde g$ at $\alpha = 1$.
\end{proof}

We also make use of the following representation of the tight IPB in terms of the tail function of the Neyman--Pearson test statistic, i.e., the likelihood ratio $L$ under $Q$.
\begin{lemma}\label{lem:ipb-representation}
  Let $g = \B^*_{\F_1}(P\|Q)$ be the tight IPB with $L = \nicefrac{dP}{dQ}\leq B$ almost surely under $Q$, and let $\alpha(t) \define \Pr_{y \sim Q}[L(y) > t]$ be the tail function of $L$ under $Q$. Then, for every $\alpha \in [0, 1]$:
 $$g(\alpha) = \int_0^B \min\{\alpha, \alpha(t)\}dt.$$
\end{lemma}
\begin{proof}
Fix $\alpha \in [0, 1]$. By \cref{stmt:ipb-np}, there exist $t_\alpha \geq 0$ and $\eta \in [0, 1]$ such that the test $\phi^* \define \phi_{t_\alpha, \eta}$ satisfies $\E_Q[\phi^*] = \alpha$ and $g(\alpha) = \E_P[\phi^*] = \E_Q[L\phi^*]$. As $0 \leq L \leq B$ almost surely under $Q$, we can write $L = \int_0^B \indic[L > t]dt$ and exchange the expectation and the integral:
$$g(\alpha) = \int_0^B \E_Q[\phi^* \indic[L > t]]dt.$$
For $t \geq t_\alpha$, we have $\phi^* \indic[L > t] = \indic[L > t]$, with expectation $\alpha(t) \leq \alpha$. For $t < t_\alpha$, we have $\phi^* \indic[L > t] = \phi^*$, with expectation $\alpha \leq \alpha(t)$. Thus, the integrand equals $\min\{\alpha, \alpha(t)\}$ for every $t$, which completes the proof.
\end{proof}

Next, we can show the main result.

\ipbnpfinite*

\begin{proof}
Omitting randomized thresholding for discrete distributions for simplicity, recall from \cref{stmt:ipb-np} that the tight IPB curve $g(\alpha)$ is parametrized by thresholds $t \ge 0$:
\begin{align*}
    \alpha(t) &\define \Pr_{y \sim Q}[L(y) > t] \\
    \beta(t) &\define \Pr_{y \sim P}[L(y) > t] = \E_{y \sim Q}[L(y) \indic[L(y) > t]]
\end{align*}
with the tight IPB $g = \B^*_{\F_1}(P\|Q)$ being the boundary traced by $(\alpha(t), \beta(t))$.

\cref{alg:ipbnp} computes points based on the sorted order of the likelihood-ratio values $L_i = L(o_i)$ where $o_i \sim Q$, such that $L_{(1)}\ge L_{(2)}\ge\cdots\ge L_{(n)}.$

For any $k\in\{1, \dots, n\}$, the algorithm computes
$$\hat\alpha_k \define \frac kn, \text{ and } \hat\beta_k \define \frac 1n \sum_{j=1}^k L_{(j)},$$
and returns the normalized values $\hat g_k = \nicefrac{\hat\beta_k}{\hat\beta_n}$.
Let $\tilde g_n$ denote the piecewise-linear interpolation of the \emph{unnormalized} points:
$$(0,0),\;(\hat\alpha_1,\hat\beta_1),\ldots,(\hat\alpha_n,\hat\beta_n).$$
This interpolation corresponds to randomized thresholding on the empirical likelihood ratios. The interpolation $\hat g_n$ of the returned points satisfies $\hat g_n = \nicefrac{\tilde g_n}{\hat\beta_n}$ with $\hat\beta_n = \tilde g_n(1)$. Now define the empirical tail function:
$$A_n(t)\define \frac1n\sum_{i=1}^n \indic[L_i>t].$$
By the Dvoretzky--Kiefer--Wolfowitz (DKW) inequality, we can write with probability at least $1-\delta$ over sampling of $Q$:
\begin{equation}\label{dkw1}
    \sup_{t\geq 0} |A_n(t) - \alpha(t)|\leq \varepsilon_n,
\end{equation}
where $\varepsilon_n\define \sqrt{\frac{1}{2n}\log\frac{2}{\delta}}$.
We now express both the population and empirical boundaries using their tail functions.  According to \cref{lem:ipb-representation}, we can write
$$g(\alpha) = \int_0^B \min\{\alpha, \alpha(t)\}dt.$$
Similarly, it can be verified that the interpolated empirical boundary satisfies
$$\tilde g_n(\alpha)=\int_0^B \min\{\alpha,A_n(t)\}\,dt.$$
Indeed, at $\alpha = \tfrac kn$, we have
$$\tilde g_n\left(\frac{k}{n}\right)=\int_0^B \min\left\{\frac{k}{n},A_n(t)\right\}dt=\frac1n\sum_{j=1}^k L_{(j)}=\hat\beta_k,$$ and the equality extends to all \(\alpha\in[0,1]\) by linear interpolation.
Therefore, for every $\alpha\in [0, 1]$, we have:
\begin{align*}
|\tilde g_n(\alpha)-g(\alpha)|
&\le\int_0^B
\left|\min\{\alpha,A_n(t)\}-\min\{\alpha,\alpha(t)\}
\right|dt \\
&\le\int_0^B |A_n(t)-\alpha(t)|\,dt \\
&\le B\varepsilon_n,
\end{align*}
where the last step follows from \eqref{dkw1}. Taking the supremum over $\alpha$, we obtain
$$\sup_{\alpha\in [0, 1]} |\tilde g_n(\alpha)-g(\alpha)|\leq B\varepsilon_n.$$
Finally, we account for the normalization. The tight IPB satisfies $g(1) = 1$, and $\tilde g_n$ is non-negative and non-decreasing, so \cref{lem:normalization} applied with $r = B\varepsilon_n$ yields:
$$\sup_{\alpha\in [0, 1]} |\hat g_n(\alpha)-g(\alpha)|\leq 2B\varepsilon_n.$$
Thus, we conclude that, with probability at least $1-\delta$,
$$\sup_{\alpha\in[0,1]}|\hat g_n(\alpha)-g(\alpha)|\le \sqrt{2}B\sqrt{\frac{\log(2/\delta)}{n}}.$$
\end{proof}

\begin{remark}
The above proof shows that under the assumption of bounded likelihood, the estimator $\hat g_n$ obtained from \cref{alg:ipbnp} converges uniformly to the tight IPB. In the unbounded case, however, such a distribution-free finite-sample guarantee does not hold in general. Instead, applying the algorithm to the truncated likelihood ratios $\min\{L, B\}$ yields an estimator $\hat g_{n, B}$ with the following bias-variance decomposition: 
\begin{equation}\label{eq:dkw-truncated}
    \sup_{\alpha\in[0,1]}|\hat g_{n,B}(\alpha)-g(\alpha)| \le C\left(B\sqrt{\frac{\log(2/\delta)}{n}}+\E_Q[(L-B)_+]\right).
\end{equation}
The first term is the statistical error for estimating the truncated boundary, whereas the second term is the truncation bias. As $P\ll Q$ implies $\E_Q[L]=1$, the truncation bias vanishes as $B\to\infty$. Thus, by choosing $B=B_n\to\infty$ slowly enough such that $B_n\sqrt{\nicefrac{\log(2/\delta)}{n}} \to 0$, the truncated estimator $\hat g_{n,B_n}$ converges uniformly to $g$ with high probability.
\end{remark}
\begin{proof}
If $L$ is unbounded, we can apply the above proof to the truncated likelihood:
$$L_{B}\define \min\{L,B\}.$$
We also modify Algorithm~\ref{alg:ipbnp} to truncate sample likelihood:
$$L_{i,B}\define \min\{L_i,B\}.$$
Define the corresponding population boundary:
$$g_B(\alpha)\define\sup_{f \in \F_1:~\E_Q[f]\le\alpha}\E_Q[L_B f].$$
Let \(\tilde g_{n,B}\) denote the piecewise-linear interpolation of the unnormalized points obtained by applying
\cref{alg:ipbnp} to the truncated values $L_{1,B},\ldots,L_{n,B}$, and let $\hat g_{n,B} = \nicefrac{\tilde g_{n,B}}{\tilde g_{n,B}(1)}$ denote the returned normalized estimator.
Applying the bounded argument (before normalization) to $L_B$, with probability at least $1-\delta$,
\begin{equation}\label{eq:dkw-unbounded}
    \sup_{\alpha\in[0,1]}|\tilde g_{n,B}(\alpha)-g_B(\alpha)| \le CB\sqrt{\frac{\log(2/\delta)}{n}}.
\end{equation}
Moreover, for every feasible $f\in\F_1$ satisfying $\E_Q[f]\leq \alpha$:
\begin{align*}
\E_Q[Lf]
&=\E_Q[L_B f]+\E_Q[(L-B)_+f] \\
&\le\E_Q[L_B f]+\E_Q[(L-B)_+] \\
&\le g_B(\alpha)+\E_Q[(L-B)_+].
\end{align*}
Taking the supremum over all feasible \(f\in \F_1\) gives
$$g(\alpha)=\sup_{f\in\F_1:~\E_Q[f]\le\alpha}\E_Q[Lf]\le g_B(\alpha)+\E_Q[(L-B)_+].$$
Also, as $L_B\leq L$, we have:
$$g_B(\alpha)\leq g(\alpha),$$
which implies: 
$$0\leq g(\alpha)-g_B(\alpha)\leq \E_Q[(L-B)_+].$$
We can write for any $\alpha\in [0,1]$:
\begin{align*}
|\tilde g_{n,B}(\alpha)-g(\alpha)|
&\le|\tilde g_{n,B}(\alpha)-g_B(\alpha)|+|g_B(\alpha)-g(\alpha)| \\
&\le|\tilde g_{n,B}(\alpha)-g_B(\alpha)| +\E_Q[(L-B)_+].
\end{align*}
Thus, we can write:
$$\sup_{\alpha\in [0, 1]}|\tilde g_{n,B}(\alpha)-g(\alpha)| \leq \sup_{\alpha\in [0, 1]} |\tilde g_{n,B}(\alpha)-g_B(\alpha)| +\E_Q[(L-B)_+].$$
Combining this with \cref{eq:dkw-unbounded} yields with probability at least $1-\delta$:
$$\sup_{\alpha\in [0, 1]}|\tilde g_{n,B}(\alpha)-g(\alpha)| \leq C B \sqrt{\frac{\log(2/\delta)}{n}} +\E_Q[(L-B)_+].$$
Finally, since $g(1) = 1$ and $\tilde g_{n,B}$ is non-negative and non-decreasing, \cref{lem:normalization} applied with $r = CB\sqrt{\nicefrac{\log(2/\delta)}{n}} + \E_Q[(L-B)_+]$ yields the sought form.
\end{proof}

\subsection{Analysis of Algorithm~\ref{alg:lagrangian}}
Let us fix the notation for this section. Throughout, $\F \subseteq \F_1$ and $\Lambda \ni 1$ denotes the grid used by \cref{alg:lagrangian}, so that $\lambda_{\max} \define \max \Lambda \geq 1$. We write:
$$\hat\E_P[f] \define \frac{1}{n_P}\sum_{x \in X_P} f(x), \qquad \hat\E_Q[f] \define \frac{1}{n_Q}\sum_{x \in X_Q} f(x),$$
to denote the empirical averages.
To distinguish the exact empirical objective from the value returned by the optimization oracle, we introduce $\hat h_\mathrm{opt}$:
$$\hat h_{\mathrm{opt}}(\lambda) \define \sup_{f \in \F}\left[\hat\E_P[f] - \lambda \hat\E_Q[f]\right], \text{ with } \left|\hat h(\lambda) - \hat h_{\mathrm{opt}}(\lambda)\right| \leq \varepsilon_{\mathrm{opt}} \ \text{ for all } \lambda \in \Lambda,$$
where $\hat h(\lambda)$ is the value used in \cref{alg:lagrangian}. The output of the algorithm is $\hat g(\alpha) = \min_{\lambda \in \Lambda}[\hat h(\lambda) + \lambda\alpha]$, and we denote by $\smash{\hat g_{\mathrm{opt}}}$ the output under an exact oracle:
$$\hat g_{\mathrm{opt}}(\alpha) \define \min_{\lambda \in \Lambda}[\hat h_{\mathrm{opt}}(\lambda) + \lambda\alpha].$$
We consider the following population \textit{dual IPBs}, each of which is a valid IPB in $\B_\F(P ~\|~ Q)$ by \cref{stmt:lagrangian}:
\begin{align*}
    g_\Lambda(\alpha) &\define \min_{\lambda \in \Lambda}\left[h_\F(\lambda) + \lambda\alpha\right] &&\text{(grid-restricted dual IPB)}, \\
    g_{[0, \lambda_{\max}]}(\alpha) &\define \inf_{0 \leq \lambda \leq \lambda_{\max}}\left[h_\F(\lambda) + \lambda\alpha\right] &&\text{(truncated dual IPB)}, \\
    g_{\mathrm{dual}}(\alpha) &\define \inf_{\lambda \geq 0}\left[h_\F(\lambda) + \lambda\alpha\right] &&\text{(Lagrangian dual IPB)}.
\end{align*}
By construction, $g(\alpha) \leq g_{\mathrm{dual}}(\alpha) \leq g_{[0, \lambda_{\max}]}(\alpha) \leq g_\Lambda(\alpha)$ for every $\alpha \in [0, 1]$, where $g = \B^*_\F(P ~\|~ Q)$ is the tight IPB.

For a distribution $S \in \{P, Q\}$ and a sample size $n$, the \emph{Rademacher complexity} of $\F$ is
$$\mathfrak{R}_{n}(S, \F) \define \E\left[\sup_{f \in \F} \frac{1}{n}\sum_{i=1}^{n} \sigma_i f(x_i)\right],$$
where the expectation is over $x_1, \ldots, x_n$ drawn i.i.d.\ from $S$ and independent Rademacher variables $\sigma_1, \ldots, \sigma_n$, each uniform on $\{-1, +1\}$. With $\varepsilon_P$ and $\varepsilon_Q$ as in \cref{stmt:lagrangian-finite}, define the event
\begin{equation}\label{eq:uniform-event}
    \mathcal{E} \define \left\{ \sup_{f \in \F}\left|\hat\E_P[f] - \E_P[f]\right| \leq \varepsilon_P \ \text{ and } \ \sup_{f \in \F}\left|\hat\E_Q[f] - \E_Q[f]\right| \leq \varepsilon_Q \right\}.
\end{equation}
As functions in our $\F$ are bounded in $[0, 1]$, standard Rademacher complexity bounds guarantee that, for each $S \in \{P, Q\}$, the upper and the lower uniform deviations of $\hat\E_S[f]$ from $\E_S[f]$ each exceed $\varepsilon_S$ with probability at most $\delta/4$. By a union bound over these four events, we have that $\Pr[\mathcal{E}] \geq 1 - \delta$.

The proof proceeds by decomposing the error of $\hat g$ with respect to $g$ into an optimization error, a statistical error, and three structural errors---grid discretization, dual-range truncation, and the duality gap:
\begin{equation}\label{eq:full-decomposition}
    \hat g - g = \underbrace{\left[\hat g - \hat g_{\mathrm{opt}}\right]}_{\text{optimization}} + \underbrace{\left[\hat g_{\mathrm{opt}} - g_\Lambda\right]}_{\text{statistical}} + \underbrace{\left[g_\Lambda - g_{[0, \lambda_{\max}]}\right]}_{\text{discretization}} + \underbrace{\left[g_{[0, \lambda_{\max}]} - g_{\mathrm{dual}}\right]}_{\text{truncation}} + \underbrace{\left[g_{\mathrm{dual}} - g\right]}_{\text{duality gap}}.
\end{equation}
We control each term in turn.

\paragraph{Optimization error.} Let us show how optimization error propagates to the IPB estimate.
\begin{lemma}\label{lem:oracle}
If $|\hat h(\lambda) - \hat h_{\mathrm{opt}}(\lambda)| \leq \varepsilon_{\mathrm{opt}}$ for all $\lambda \in \Lambda$, then $\sup_{\alpha \in [0, 1]}|\hat g(\alpha) - \hat g_{\mathrm{opt}}(\alpha)| \leq \varepsilon_{\mathrm{opt}}$.
\end{lemma}
\begin{proof}
Using $|\min_x A(x) - \min_x B(x)| \leq \max_x |A(x) - B(x)|$, for any $\alpha$ we have $|\hat g(\alpha) - \hat g_{\mathrm{opt}}(\alpha)| \leq \max_{\lambda \in \Lambda}|\hat h(\lambda) - \hat h_{\mathrm{opt}}(\lambda)| \leq \varepsilon_{\mathrm{opt}}$.
\end{proof}

\paragraph{Statistical error.}
We first show that, on the event $\mathcal{E}$, the empirical Lagrangian objective is uniformly close to the population one.
\begin{lemma}\label{lem:linewise}
On the event $\mathcal{E}$, simultaneously for every $\lambda \geq 0$,
$$\left|\hat h_{\mathrm{opt}}(\lambda) - h_\F(\lambda)\right| \leq \varepsilon_P + \lambda\varepsilon_Q.$$
\end{lemma}
\begin{proof}
By the triangle inequality and the subadditivity of the supremum,
$$\left|\hat h_{\mathrm{opt}}(\lambda) - h_\F(\lambda)\right| \leq \sup_{f \in \F}\left|\hat\E_P[f] - \E_P[f]\right| + \lambda\sup_{f \in \F}\left|\hat\E_Q[f] - \E_Q[f]\right| \leq \varepsilon_P + \lambda\varepsilon_Q,$$
where the last step holds on the event $\mathcal{E}$. As $\mathcal{E}$ does not depend on $\lambda$, the bound holds for all $\lambda \geq 0$ simultaneously.
\end{proof}
The statistical error at a given $\alpha$ is defined by the values of $\lambda$ that are actually solutions to the optimization in \cref{alg:lagrangian} at $\alpha$. For a fixed $\alpha \in [0, 1]$, define the \emph{active} population and empirical dual variables as follows:
$$\lambda^*_\alpha \in \arg\min_{\lambda \in \Lambda}\left[h_\F(\lambda) + \lambda\alpha\right], \qquad \hat\lambda_\alpha \in \arg\min_{\lambda \in \Lambda}\left[\hat h_{\mathrm{opt}}(\lambda) + \lambda\alpha\right].$$
\begin{proposition}\label{stmt:pointwise-statistical}
Suppose $\F \subseteq \F_1$. On the event $\mathcal{E}$, for every $\alpha \in [0, 1]$,
$$\left|\hat g_{\mathrm{opt}}(\alpha) - g_\Lambda(\alpha)\right| \leq \varepsilon_P + \max\left\{\lambda^*_\alpha, \hat\lambda_\alpha\right\}\varepsilon_Q \leq \varepsilon_P + \lambda_{\max}\varepsilon_Q.$$
\end{proposition}
\begin{proof}
Fix $\alpha$. Using $\hat g_{\mathrm{opt}}(\alpha) \leq \hat h_{\mathrm{opt}}(\lambda^*_\alpha) + \lambda^*_\alpha\alpha$ and \cref{lem:linewise},
$$\hat g_{\mathrm{opt}}(\alpha) - g_\Lambda(\alpha) \leq \hat h_{\mathrm{opt}}(\lambda^*_\alpha) - h_\F(\lambda^*_\alpha) \leq \varepsilon_P + \lambda^*_\alpha\varepsilon_Q.$$
Symmetrically, using $g_\Lambda(\alpha) \leq h_\F(\hat\lambda_\alpha) + \hat\lambda_\alpha\alpha$,
$$g_\Lambda(\alpha) - \hat g_{\mathrm{opt}}(\alpha) \leq h_\F(\hat\lambda_\alpha) - \hat h_{\mathrm{opt}}(\hat\lambda_\alpha) \leq \varepsilon_P + \hat\lambda_\alpha\varepsilon_Q.$$
Combining the two directions yields the first inequality; the second follows from $\max\{\lambda^*_\alpha, \hat\lambda_\alpha\} \leq \lambda_{\max}$.
\end{proof}

\paragraph{Grid discretization.}
Next, we show the effect of the chosen grid density. First, we show that the dual objective is Lipschitz in $\lambda$.
\begin{lemma}\label{lem:lipschitz}
For $\F \subseteq \F_1$ and every fixed $\alpha \in [0, 1]$, the map $\lambda \mapsto h_\F(\lambda) + \lambda\alpha$ is $1$-Lipschitz on $[0, \infty)$.
\end{lemma}
\begin{proof}
For each $f \in \F$ and any $\lambda, \lambda' \geq 0$, we have:
\begin{align*}
\left|\left(\E_P[f] - \lambda\E_Q[f] + \lambda\alpha\right) - \left(\E_P[f] - \lambda'\E_Q[f] + \lambda'\alpha\right)\right|
&= \left|\alpha - \E_Q[f]\right| \cdot \left|\lambda - \lambda'\right| \\
&\leq \left|\lambda - \lambda'\right|,
\end{align*}
where the last step holds as $\alpha$ and $\E_Q[f] \in [0, 1]$. A pointwise supremum of $1$-Lipschitz maps is $1$-Lipschitz.
\end{proof}
As a consequence, replacing the minimizer over $[0, \lambda_{\max}]$ by its nearest grid point costs at most the covering radius of the grid.
\begin{proposition}\label{stmt:discretization}
Suppose $\F \subseteq \F_1$. Then, for every $\alpha \in [0, 1]$,
$$0 \leq g_\Lambda(\alpha) - g_{[0, \lambda_{\max}]}(\alpha) \leq \eta_\Lambda.$$
In particular, if $\Lambda$ is a uniform grid of $k = |\Lambda|$ points on $[0, \lambda_{\max}]$, then $\eta_\Lambda = \frac{\lambda_{\max}}{2(k-1)}$, and to make the discretization error at most $\varepsilon$ it suffices to take $k \geq 1 + \frac{\lambda_{\max}}{2\varepsilon}$.
\end{proposition}
\begin{proof}
The lower bound holds as $\Lambda \subseteq [0, \lambda_{\max}]$. For the upper bound, fix $\alpha$. By \cref{lem:lipschitz}, the map $\lambda \mapsto h_\F(\lambda) + \lambda\alpha$ is continuous, so it attains its minimum over $[0, \lambda_{\max}]$ at some $\lambda^*$. Let $\lambda' \in \Lambda$ be the nearest grid point, so that $|\lambda^* - \lambda'| \leq \eta_\Lambda$. By \cref{lem:lipschitz} again,
$$g_\Lambda(\alpha) \leq h_\F(\lambda') + \lambda'\alpha \leq h_\F(\lambda^*) + \lambda^*\alpha + \eta_\Lambda = g_{[0, \lambda_{\max}]}(\alpha) + \eta_\Lambda.$$
The expression for a uniform grid is immediate.
\end{proof}

\paragraph{Dual-range truncation.}
Next, we analyze the error due to truncating the dual variable to $[0, \lambda_{\max}]$. It is zero if the continuous dual problem has a minimizer in $[0, \lambda_{\max}]$. Otherwise, the minimizing dual variable lies beyond $\lambda_{\max}$, and no refinement of the grid inside $[0, \lambda_{\max}]$ can remove this error. First, we show that under a bounded likelihood ratio, the Lagrangian objective vanishes for all large enough dual variables.
\begin{lemma}\label{stmt:h-vanishes}
Assume $0 \in \F \subseteq \F_1$, $P \ll Q$, and $\nicefrac{dP}{dQ} \leq B$ almost surely under $Q$ for some $B < \infty$. Then, for every $\lambda \geq B$, we have $h_\F(\lambda) = 0$.
\end{lemma}
\begin{proof}
As $0 \in \F$, the zero function gives $h_\F(\lambda) \geq 0$. For any $f \in \F \subseteq \F_1$, as $P \ll Q$ and $0 \leq f \leq 1$, we have:
$$\E_P[f] - \lambda\E_Q[f] = \E_Q\big[(\nicefrac{dP}{dQ} - \lambda)f\big] \leq \E_Q\big[(\nicefrac{dP}{dQ} - \lambda)_+\big].$$
Taking the supremum over $f \in \F$ gives $h_\F(\lambda) \leq \E_Q\big[(\nicefrac{dP}{dQ} - \lambda)_+\big]$. For $\lambda \geq B$, we have $\nicefrac{dP}{dQ} - \lambda \leq 0$ almost surely under $Q$, so the right-hand side is zero, and hence $h_\F(\lambda) = 0$.
\end{proof}
As a consequence, the truncation error vanishes for the principled choice $\lambda_{\max} \geq B$.
\begin{proposition}\label{stmt:truncation}
Assume $0 \in \F \subseteq \F_1$, $P \ll Q$, and $\nicefrac{dP}{dQ} \leq B$ almost surely under $Q$ for some $B < \infty$. If $\lambda_{\max} \geq B$, then $g_{[0, \lambda_{\max}]}(\alpha) = g_{\mathrm{dual}}(\alpha)$ for every $\alpha \in [0, 1]$.
\end{proposition}
\begin{proof}
By \cref{stmt:h-vanishes}, $h_\F(\lambda) = 0$ for all $\lambda \geq B$. Hence, for any $\lambda \geq B$ and $\alpha \geq 0$, we have $h_\F(\lambda) + \lambda\alpha = \lambda\alpha \geq B\alpha = h_\F(B) + B\alpha$, so no dual variable larger than $B$ can improve the objective. Therefore, the infimum over $\lambda \geq 0$ coincides with the infimum over $[0, B] \subseteq [0, \lambda_{\max}]$.
\end{proof}

\paragraph{Duality gap.}
Finally, we analyze the last term that vanishes when we have strong duality:
\begin{proposition}[Strong duality]\label{stmt:strong-duality}
Suppose $0 \in \F \subseteq \F_1$ is convex, $P \ll Q$, and $\nicefrac{dP}{dQ} \leq B$ almost surely under $Q$ for some $B < \infty$. Then $g_{\mathrm{dual}}(\alpha) = g(\alpha)$ for every $\alpha \in [0, 1]$.
\end{proposition}

\begin{proof}
Let $g = \B^*_{\F}(P \| Q)$ be the tight IPB. Since $\F$ is convex, $g$ is concave. Moreover,  since $0\in\F\subseteq\F_1$, it is finite and non-decreasing on $[0,1]$. Thus, for every $\alpha\in(0,1)$, $g$ admits a supergradient $\lambda_\alpha\geq0$ by \citet[Theorem~23.4]{rockafellar1997convex}, so that:
$$g(t)\leq g(\alpha)+\lambda_\alpha(t-\alpha),\qquad t\in[0,1].$$
In particular, for every $f\in\F$,
$$\E_P[f]\leq g(\E_Q[f])\leq g(\alpha)+\lambda_\alpha(\E_Q[f]-\alpha).$$
Taking the supremum over $f\in\F$ yields $h_\F(\lambda_\alpha)+\lambda_\alpha\alpha\leq g(\alpha)$. The reverse inequality follows from \cref{stmt:lagrangian}, establishing strong duality.
At $\alpha=1$, equality follows by choosing $\lambda=0$, since $h_\F(0)=g(1)$. At $\alpha=0$, the bounded likelihood-ratio assumption gives $h_\F(B)=0$ by \cref{stmt:h-vanishes}, whereas $g(0)=0$ since $P\ll Q$ and all functions in $\F$ are non-negative. As a result, strong duality holds throughout $[0,1]$.
\end{proof}

\paragraph{Main result.} We are now ready to prove the main statement.

\lagrangianfinite*
\begin{proof}
On the event $\mathcal{E}$ in \cref{eq:uniform-event}, which holds with probability at least $1 - \delta$, we bound each term of the decomposition in \cref{eq:full-decomposition} uniformly in $\alpha \in [0, 1]$. The optimization term is at most $\varepsilon_{\mathrm{opt}}$ in absolute value by \cref{lem:oracle}. The statistical term is at most $\varepsilon_P + \lambda_{\max}\varepsilon_Q$ in absolute value by \cref{stmt:pointwise-statistical}. The discretization term lies in $[0, \eta_\Lambda]$ by \cref{stmt:discretization}. The truncation term is zero by \cref{stmt:truncation}, as $\lambda_{\max} \geq B$. The duality gap is zero by \cref{stmt:strong-duality}. The claim follows from the triangle inequality.
\end{proof}

\paragraph{Sharpening the dependence on the dual variable.}
The dependence on $\lambda_{\max}$ in \cref{stmt:lagrangian-finite} is conservative. By \cref{stmt:pointwise-statistical}, for every fixed $\alpha \in [0, 1]$, the term $\lambda_{\max}\varepsilon_Q$ can be replaced by $\smash{\max\{\lambda^*_\alpha, \hat\lambda_\alpha\}}\varepsilon_Q$, where $\lambda^*_\alpha$ and $\smash{\hat\lambda_\alpha}$ are the active population and empirical dual variables at $\alpha$. This yields a pointwise bound that is tighter whenever the active dual variable at $\alpha$ is below $\lambda_{\max}$.

\paragraph{Extension to non-convex function families.}
Dropping the convexity assumption on $\F$, the same argument results in the following bound:
$$\sup_{\alpha \in [0, 1]}\left|\hat g(\alpha) - g(\alpha)\right| \leq \varepsilon_P + \lambda_{\max}\varepsilon_Q + \eta_\Lambda + \varepsilon_{\mathrm{opt}} + \sup_{\alpha \in [0, 1]}\left[g_{\mathrm{dual}}(\alpha) - g(\alpha)\right],$$
where the last term is the duality gap. For a general non-convex learnable family $\F$, strong duality need not hold, and thus $g_{\mathrm{dual}} - g$ is a structural gap: the other remaining terms do not control it, as grid density and $\lambda_{\max}$ only control the finite-grid approximation to the truncated dual problem. In this setting, the output of \cref{alg:lagrangian} should be interpreted as producing the Lagrangian dual IPB $g_{\mathrm{dual}}$ rather than the tight IPB, which by \cref{stmt:lagrangian} remains a valid IPB in $\B_\F(P ~\|~ Q)$.

\paragraph{Extension to unbounded function families.}
\cref{alg:lagrangian} itself is applicable to functions with an arbitrary range, although we state  \cref{stmt:lagrangian-finite} for $\F \subseteq \F_1$. The guarantee extends directly to a bounded range $[-M, M]$ by scaling: we multiply the statistical term $\varepsilon_S$ for $S \in \{P, Q\}$ and the Lipschitz constant in \cref{lem:lipschitz}---hence the discretization error in \cref{stmt:discretization}---by $2M$. 

For unbounded families, we use a clipping argument analogous to \cref{eq:dkw-truncated}. Define the clipping operator and the clipped family:
$$T_M(u) \define \max\{-M, \min\{u, M\}\}, \qquad \F_M \define \{T_M \circ f \mid f \in \F\},$$
and run \cref{alg:lagrangian} on $\F_M$, i.e., with the empirical objective
$$\hat h_{M, \mathrm{opt}}(\lambda) \define \sup_{f \in \F}\left[\hat\E_P[T_M(f)] - \lambda\hat\E_Q[T_M(f)]\right],$$
which is exactly \cref{eq:algo-opt-lagrangian} with $\F$ replaced by its clipped version $\F_M$. As before, we denote by $\hat h_M(\lambda)$ the value returned by the optimization oracle, with $|\hat h_M(\lambda) - \hat h_{M, \mathrm{opt}}(\lambda)| \leq \varepsilon_{\mathrm{opt}}$ for all $\lambda \in \Lambda$. The resulting IPB estimate is $\hat g_{\Lambda, M}(\alpha) \define \min_{\lambda \in \Lambda}[\hat h_M(\lambda) + \lambda\alpha]$. The only extra error is the bias due to clipping. As $|f - T_M(f)| = (|f| - M)_+$, we can quantify it by:
$$\tau_{S, M} \define \sup_{f \in \F}\E_S\left[(|f| - M)_+\right], \qquad S \in \{P, Q\}.$$

\begin{proposition}\label{stmt:clipping}
Let $\F$ be a family of real-valued functions with $\tau_{P, M}, \tau_{Q, M} < \infty$. Then, for every $\lambda \geq 0$,
$$\left|h_\F(\lambda) - h_{\F_M}(\lambda)\right| \leq \tau_{P, M} + \lambda\tau_{Q, M}.$$
Consequently, with probability at least $1 - \delta$, the output $\hat g_{\Lambda, M}$ of \cref{alg:lagrangian} run on $\F_M$ with an optimization oracle of error at most $\varepsilon_{\mathrm{opt}}$ satisfies
$$\sup_{\alpha}\left|\hat g_{\Lambda, M}(\alpha) - g_\Lambda(\alpha)\right| \leq \varepsilon_{P, M} + \lambda_{\max}\varepsilon_{Q, M} + \tau_{P, M} + \lambda_{\max}\tau_{Q, M} + \varepsilon_{\mathrm{opt}},$$
where $\varepsilon_{S, M} \define 2\mathfrak{R}_{n_S}(S, \F_M) + M\sqrt{\nicefrac{2\log(4/\delta)}{n_S}}$ for $S \in \{P, Q\}$, and $g_\Lambda$ is the grid-restricted dual IPB for the original family $\F$.
\end{proposition}

\begin{proof}
For the second claim, let $\hat g_{\Lambda, M, \mathrm{opt}}(\alpha) \define \min_{\lambda \in \Lambda}[\hat h_{M, \mathrm{opt}}(\lambda) + \lambda\alpha]$ denote the estimate under an exact oracle. By \cref{lem:oracle}, $\sup_\alpha|\hat g_{\Lambda, M}(\alpha) - \hat g_{\Lambda, M, \mathrm{opt}}(\alpha)| \leq \varepsilon_{\mathrm{opt}}$.
Since $\F_M$ is a family of $[-M,M]$-valued functions, symmetrization implies:
$$\E\left[\sup_{\varphi\in\F_M}\left[\hat\E_S[\varphi]-\E_S[\varphi]\right]\right]\leq 2\mathfrak R_{n_S}(S,\F_M).$$
Also, since $|\varphi|\leq M$, replacing one observation changes the empirical average of any $\varphi\in\F_M$ by at most $2M/n_S$. The same bounded-differences property holds for the supremum. The analogous argument applies to the negative deviation $\sup_{\varphi\in\F_M} \big[ \E_S[\varphi]-\hat\E_S[\varphi] \big]$, using the symmetry of the Rademacher variables. Thus, McDiarmid's inequality and a union bound over both deviations and $S\in\{P,Q\}$ yield, with probability at least $1-\delta$ for any $\delta \in (0, 1)$,
$$\sup_{\varphi\in\F_M}|\hat\E_S[\varphi]-\E_S[\varphi]|\leq\varepsilon_{S,M},\qquad S\in\{P,Q\},$$
Thus, the same argument as in \cref{stmt:pointwise-statistical} gives:
$$\sup_\alpha|\hat g_{\Lambda, M, \mathrm{opt}}(\alpha) - g_{\Lambda, M}(\alpha)| \leq \varepsilon_{P, M} + \lambda_{\max}\varepsilon_{Q, M},$$
where $g_{\Lambda, M}(\alpha) \define \min_{\lambda \in \Lambda}[h_{\F_M}(\lambda) + \lambda\alpha]$.
By the first claim and $|\min_x A(x) - \min_x B(x)| \leq \max_x|A(x) - B(x)|$, we have $\sup_\alpha|g_{\Lambda, M}(\alpha) - g_\Lambda(\alpha)| \leq \tau_{P, M} + \lambda_{\max}\tau_{Q, M}$. The triangle inequality completes the proof. \\
\end{proof}

This extension demonstrates a trade-off. Increasing $M$ makes the clipped family closer to the original one and reduces the tail bias $\tau_{S, M}$, but increases the bounded-range error term. The bias terms vanish fast under mild tail assumptions: e.g., if $f(x)$ is $\sigma$-sub-Gaussian under $S$ uniformly over $f \in \F$, then $\tau_{S, M} = \mathcal{O}\left(\sigma \exp(\nicefrac{-M^2}{2\sigma^2})\right)$, so choosing $M \asymp \sigma\sqrt{\log n}$ yields a total error of order $\varepsilon_{P} + \lambda_{\max}\varepsilon_Q$ up to logarithmic factors. 

\subsection{Proofs for Section~\ref{sec:applications}}

\bonipb*
\begin{proof}
The proof for the case without ties is given in the main body. The no-ties assumption in \cref{stmt:bon-ipb} is only needed to obtain the exact closed form $g(\alpha)=1-(1-\alpha)^N$. Without this assumption, the same expression remains a valid, though not necessarily tight, IPB.
Indeed, let $Y_1,\ldots,Y_N \sim Q$, and let $\smash{\hat Y}$ be the output selected by
BoN, with any tie-breaking rule that selects one of the maximizers uniformly at random. For any $f\in\F_1$,
$$f(\hat Y)\le\max_{i\in[N]} f(Y_i)\le1-\prod_{i=1}^N (1-f(Y_i)).$$

To see why the second inequality holds, let $i^*$ be an index attaining the maximum. As $0 \leq f \leq 1$, every factor $1 - f(Y_i)$ lies in $[0, 1]$, so $\prod_{i=1}^N (1-f(Y_i)) \leq 1 - f(Y_{i^*})$, thus we get the inequality by re-arranging. Taking expectations and using independence gives:
$$\E_{P_N}[f]\le1-\prod_{i=1}^N \E_Q[1-f(Y_i)]=1-\left(1-\E_Q[f]\right)^N .$$
Therefore, for every $\alpha\in[0,1]$,
$$g(\alpha)=\sup_{f\in\F_1:\,\E_Q[f]\le\alpha}\E_{P_N}[f]\le1-(1-\alpha)^N.$$
\end{proof}

\section{Additional Results}
\label{app:extra-results}

\subsection{IPBs and $f$-Divergences}
We can obtain an IPB using a standard information-theoretic notion of an $f$-divergence. We call them $\phi$-divergences in this manuscript to avoid the clash of nomenclature with $f \in \F$.

\SA{Equation (2.79) in Polyanskiy's thesis can also be useful to connect IPB to $\phi$-divergence. It implies the following: If $P\ll Q$ and the distribution of $dP/dQ$ does not have any positive mass, then $$D_\phi(P\|Q)=\int\phi(L)\,dQ=\int_0^1\phi\!\left((g^*_{P,Q})'(\alpha)\right)d\alpha,$$
showing the tight IPB fully characterize $\phi$-divergence.}

\begin{definition}[$\phi$-divergence]
Let $P$ and $Q$ be probability distributions over a space $\mathcal{X}$ such that $P$ is absolutely continuous with respect to $Q$ ($P \ll Q$). Let $\phi: (0, \infty) \to \mathbb{R}$ be a convex function such that $\phi(1) = 0$. The $\phi$-divergence of $P$ from $Q$ is defined as:
\begin{equation}
    D_\phi(P \| Q) = \mathbb{E}_Q \left[ \phi\left( \frac{dP}{dQ} \right) \right]
\end{equation}
\end{definition}

\begin{restatable}[$\phi$-divergence to IPB]{proposition}{fdiv}
  For any valid $\phi$-divergence $D_\phi(P \| Q)$, $g(\alpha)$ as defined next is an IPB, $g \in \B_{\F_1}(P \| Q)$:
  \begin{equation}
      g(\alpha) = \inf_{t > 0} \left\{ \frac{D_\phi(P \| Q) + \alpha \phi^*(t) + (1 - \alpha) \phi^*(0)}{t} \right\}
  \end{equation}
  where $\phi^*(y) \define \sup_{x \in \R} (xy - \phi(x))$ is the convex conjugate of $\phi$.
\end{restatable}

\begin{proof}
By the variational representation of $\phi$-divergences:
\begin{equation*}
  D_\phi(P \| Q) = \sup_{h:\ \X \to \R} \{ \E_P[h] - \E_Q[\phi^*(h)] \}.
\end{equation*}
Fix any $f \in \F_1$ with $\E_Q[f] = \alpha$, and let $\beta \define \E_P[f]$. For any $t > 0$, taking $h(x) = t f(x)$ in the variational representation yields a lower bound on the divergence:
\begin{equation*}
  D_\phi(P \| Q) \ge \E_P[tf] - \E_Q[\phi^*(tf)] = t\beta - \E_Q[\phi^*(tf)].
\end{equation*}

Because the range of $f$ is bounded within $[0, 1]$, we can bound the convex conjugate term using Jensen's inequality. For any fixed $x$, we treat $f(x)$ as the expectation of a Bernoulli random variable $Z_x
\sim \mathsf{Bern}(f(x))$. As $\phi^*$ is convex, by Jensen's inequality:
\begin{equation*}
  \phi^*(tf(x)) = \phi^*(\E[tZ_x]) \le \E[\phi^*(tZ_x)] = f(x)\phi^*(t) + (1-f(x))\phi^*(0).
\end{equation*}
Taking the expectation under $Q$ and using $\E_Q[f] = \alpha$:
\begin{equation*}
  \E_Q[\phi^*(tf)] \le \alpha \phi^*(t) + (1 - \alpha)\phi^*(0).
\end{equation*}
Combining with the variational inequality and rearranging to isolate $t\beta$:
\begin{equation*}
  t\beta \le D_\phi(P \| Q) + \alpha \phi^*(t) + (1 - \alpha) \phi^*(0).
\end{equation*}
Dividing by $t > 0$ and taking the infimum over all $t > 0$:
\begin{equation*}
  \beta \;\le\; \inf_{t > 0} \left\{ \frac{D_\phi(P \| Q) + \alpha \phi^*(t) + (1 - \alpha) \phi^*(0)}{t} \right\} \;=\; g(\alpha).
\end{equation*}
This holds for any $f \in \F_1$ with $\E_Q[f] = \alpha$, so $g \in \B_{\F_1}(P ~\|~ Q)$.
\end{proof}

We apply this to KL divergence for an illustration.
\begin{corollary}
\label{stmt:kl-to-ipb}
The following function is an IPB, $g \in \B_{\F_1}(P ~\|~ Q)$:
\begin{equation}
    g(\alpha) = \inf_{t > 0} \left\{ \frac{D_{\mathsf{KL}}(P \| Q) + \alpha (e^t - 1)}{t} \right\}.
\end{equation}
\end{corollary}

\begin{proof}
We apply the general $\phi$-divergence bound from the previous theorem using $\phi(x) = x \log x - x + 1$, which generates the KL divergence, $D_{\mathsf{KL}}(P \| Q)$. Its convex conjugate
is:
\begin{equation*}
  \phi^*(y) = \sup_{x > 0} \{xy - (x \log x - x + 1)\} = e^y - 1,
\end{equation*}
giving $\phi^*(t) = e^t - 1$ and $\phi^*(0) = 0$. Substituting into the inf expression from the previous theorem:
\begin{equation*}
  g(\alpha) = \inf_{t > 0} \left\{ \frac{D_{\mathsf{KL}}(P \| Q) + \alpha (e^t - 1) + (1 - \alpha)(0)}{t} \right\} = \inf_{t > 0} \left\{ \frac{D_{\mathsf{KL}}(P \| Q) + \alpha (e^t - 1)}{t} \right\}.
\qedhere
\end{equation*}
\end{proof}
  
\subsection{Additional Algorithms}

\paragraph{Searching for the $\lambda_{\max}$ in Algorithm~\ref{alg:lagrangian}.} It is not immediately clear how to choose the grid $\Lambda$ in \cref{alg:lagrangian} in a principled way. In particular, although any $\lambda \geq 0$ is admissible, it is unclear what is a reasonable value of $\lambda_{\max}$ to set in practice. As we have shown in \cref{stmt:h-vanishes}, for regular enough $P$ and $Q$, the Lagrangian objective in \cref{eq:algo-opt-lagrangian} is zero when $\lambda$ becomes large enough. As a practical consequence, we want to heuristically find $\lambda_{\max}$ at which $\smash{\hat{h}}(\lambda_{\max})$ is numerically small enough. In \cref{alg:lambda-search} we provide a procedure to do so using doubling search until either $\smash{\hat{h}}$ is small enough, or, to prevent looping forever, until the decrease in $\smash{\hat{h}}$ becomes stagnant.

\begin{algorithm}[t]
\caption{Doubling search for $\lambda_{\max}$}\label{alg:lambda-search}
\KwIn{Samples $X_P \sim P^{n_P}$, $X_Q \sim Q^{n_Q}$; family $\F$; tolerance $\tau > 0$; target $h_{\min} > 0$}
\KwOut{Upper bound $\lambda_{\max}$}
$\lambda \gets 1$\\
\Repeat{$\hat{h}(\lambda) < h_{\min}$ \text{ or } $\hat{h}(\lambda/2) - \hat{h}(\lambda) < \tau$}{
    $\lambda \gets 2 \cdot \lambda$ \\
    $
        \hat{h}(\lambda) \gets \max_{f \in \F} \left[ \frac{1}{n_P}\sum_{x \in X_P} f(x) - \lambda \frac{1}{n_Q} \sum_{x \in X_Q} f(x) \right]
    $
}
\Return $\lambda$
\end{algorithm}

\paragraph{Estimating the lower boundary in Algorithm~\ref{alg:lagrangian} setup.}\label{app:lagrangian-lower} 

\cref{alg:lagrangian} estimates the IPB that corresponds to the upper boundary of the achievable region $\C_\F(P ~\|~ Q)$. Although we could apply the same algorithm to estimate the adjoint $\check{g} \in \B_\F(Q ~\|~ P)$ and use the generalized inverse $\check{g}^{-1}$ to get the lower boundary, for families $\F$ of, e.g., classifiers, it would be more convenient to find an algorithm that estimates the lower boundary directly. To simplify this, we present a version of the procedure that is adapted so that we do not need to compute the inverse. The adapted algorithm follows the same construction but replaces maximization with minimization, and vice versa, based on the following analogous result.
\begin{proposition}[Lagrangian dual lower boundary]\label{stmt:lagrangian-lower}
For any $\lambda \geq 0$, define
$$v_\F(\lambda) \define \inf_{f \in \F}\bigl[\E_P[f] - \lambda\,\E_Q[f]\bigr].$$
Then for every $f \in \F$ and every $\alpha \leq \E_Q[f]$,
$$
    \E_P[f] \;\geq\; \sup_{\lambda \geq 0} [v_\F(\lambda) + \lambda\,\alpha].
$$
\end{proposition}
\begin{proof}
By definition, $v_\F(\lambda) \leq \E_P[f] - \lambda \E_Q[f]$ for any $f \in \F$. Rearranging gives $\E_P[f] \geq v_\F(\lambda) + \lambda \E_Q[f]$. We get the statement by considering $\alpha \leq \E_Q[f]$.
\end{proof}

\begin{algorithm}[t]
\caption{Lower-boundary IPB estimation for a learnable family $\F$}\label{alg:lagrangian-lower}
\KwIn{Samples $X_P \sim P^{n_P}$, $X_Q \sim Q^{n_Q}$; family $\F$; grid $\Lambda = \{\lambda_1, \lambda_2, \ldots, \lambda_k\}$}
\KwOut{Lower boundary estimate $\check g^{-1}$}
\For{$\lambda \in \Lambda \cup \{1\}$}{
    \vspace{-.5em}
    \begin{equation}\label{eq:algo-opt-lagrangian-lower}
        \hat v(\lambda) \gets \min_{f \in \F} \left[ \frac{1}{n_P}\sum_{x \in X_P} f(x) - \lambda \frac{1}{n_Q} \sum_{x \in X_Q} f(x) \right]
    \end{equation}\\[-.75em]
}
$\check{g}^{-1}(\alpha) \gets \displaystyle\max_{\lambda\in\Lambda \cup \{1\}} \left[\hat{v}(\lambda)+\lambda\,\alpha\right]$\\
\Return $\check g^{-1}$.
\end{algorithm}

The finite-sample uniform-convergence guarantee of \cref{stmt:lagrangian-finite} applies with trivial changes.

\subsection{Deriving Tight Bounds on Generalization and Attack Risk in Differential Privacy}
\label{app:f-dp}

We show how IPBs unify and recover recent results on generalization and operational risk under differential privacy.
Differential privacy (DP)~\cite{dwork2014algorithmic} is the standard framework for ensuring strong privacy guarantees in statistics and machine learning. We formalize a learning algorithm as a randomized \emph{mechanism} $M : 2^\X \to \Theta$. We say that two datasets $S, S' \in \X^n$ are \emph{neighbouring}, denoted $S \simeq S'$, if they differ in at most one record (e.g., by replacement, or by addition or removal of a single record). 
Next, we introduce a form of differential privacy known as $f$-DP~\cite{dong2022gaussian}. This definition characterizes the privacy guarantees via the ROC curve of the optimal hypothesis test that distinguishes neighbouring datasets given the mechanism's output: 

\begin{definition}[\citealp{dong2022gaussian}]
    A mechanism $M: 2^{\X} \to \Theta$ satisfies $f$-DP, where $f: [0, 1] \to [0, 1]$ is a non-increasing, convex function bounded as $f(\alpha) \leq 1 - \alpha$ for $\alpha \in [0, 1]$, if for any $S \simeq S'$ and any $\alpha \in [0, 1]$, the function $f$ is a lower bound on the error rate of the most powerful hypothesis test between the distributions of $M(S)$ and $M(S')$:
    \begin{equation}
        f(\alpha) \leq \inf_{\phi:~\Theta \to [0, 1]} \{ \beta_\phi \mid \alpha_\phi \leq \alpha \},
    \end{equation}
    where $\beta_\phi \define 1 - \E_{M(S')}[\phi]$ is the false-negative rate (FNR) of the test, and $\alpha_\phi \define \E_{M(S)}[\phi]$ is the false-positive rate (FPR).
\end{definition}

By \cref{stmt:ipb-np}, we can immediately see that the function $1 - f$ is an IPB over $\F_1$:
\begin{proposition}[$f$-DP to IPB]\label{stmt:fdp-ipb}
    If a mechanism $M(\cdot)$ satisfies $f$-DP, then for any $S \simeq S'$, we have that
    $g = 1 - f$ is an IPB, $g \in \B_{\F_1}(M(S) ~\|~ M(S'))$.
\end{proposition}

Next, we recover two results recently shown by a subset of the authors~\cite{kulynych2025unifying} on generalization and risk bounds in $f$-DP.

\paragraph{Bounds on average generalization error.}
For a bounded loss $\ell: \X \times \Theta \to [0, 1]$ and a data distribution $D$ over $\X$, define the on-average train and test losses:
$$\mathrm{err}_{\mathrm{tr}} \define \E_{\substack{S \sim D^n \\ \theta \sim M(S)}}\!\left[\frac{1}{n}\sum_{i=1}^n \ell(z_i; \theta)\right], \qquad \mathrm{err}_{\mathrm{test}} \define \E_{\substack{S \sim D^n,\, z \sim D \\ \theta \sim M(S)}}[\ell(z; \theta)],$$
where $S = (z_1, \ldots, z_n)$.
\begin{proposition}[On-average generalization for $f$-DP]\label{stmt:fdp-generalization}
    Suppose that $M(\cdot)$ satisfies $f$-DP w.r.t.\ the replace-one relation, and let $g \define 1 - f$. Then, for any bounded loss function $\ell : \X \times \Theta \to [0, 1]$ and any data distribution $D$ over $\X$, we have:
    $$\mathrm{err}_{\mathrm{tr}} \leq g(\mathrm{err}_{\mathrm{test}}), \qquad \mathrm{err}_{\mathrm{test}} \leq g(\mathrm{err}_{\mathrm{tr}}).$$
\end{proposition}
\begin{proof}
    Fix any $z, z' \in \X$ and $\bar S \in \X^{n-1}$. The datasets $\bar S \cup \{z\}$ and $\bar S \cup \{z'\}$ differ in at most one record and are therefore neighbouring under the replace-one relation, so by \cref{stmt:fdp-ipb}, $g \in \B_{\F_1}(M(\bar S \cup \{z\}) ~\|~ M(\bar S \cup \{z'\}))$. Applying the main IPB property in \cref{stmt:ipb-bound} to the bounded function $\psi(\theta) \define \ell(z; \theta)$:
    $$\E_{\theta \sim M(\bar S \cup \{z\})}[\ell(z; \theta)] \;\leq\; g\!\left(\E_{\theta \sim M(\bar S \cup \{z'\})}[\ell(z; \theta)]\right).$$
    Next, we take the expectation over $z, z' \sim D$ and $\bar S \sim D^{n-1}$ on both sides. By the concavity of $g$ (\cref{stmt:ipb-concave}) and Jensen's inequality, we can pull the outer expectation inside $g(\cdot)$ on the right-hand side:
    $$\E_{\substack{z,\, \bar S \\ \theta \sim M(\bar S \cup \{z\})}}[\ell(z; \theta)] \;\leq\; g\!\left(\E_{\substack{z, z',\, \bar S \\ \theta \sim M(\bar S \cup \{z'\})}}[\ell(z; \theta)]\right).$$
    The l.h.s.\ equals $\mathrm{err}_{\mathrm{tr}}$ as the random dataset $\bar S \cup \{z\}$ is distributed as $D^n$ with $z$ being one of its records. On the r.h.s., $\bar S \cup \{z'\} \sim D^n$, but $z \sim D$ is independent of it, so the argument equals $\mathrm{err}_{\mathrm{test}}$. The reverse direction follows by symmetry, swapping the roles of $z$ and $z'$.
\end{proof}

\paragraph{Bounds on operational attack risks.}
IPBs also yield bounds on the success of re-identification, attribute inference, and data reconstruction attacks, equivalent to those recently derived by a subset of the authors~\cite{kulynych2025unifying}. The authors have also shown that these bounds are empirically the tightest known bounds on attack risk in many settings, e.g., for general reconstruction attacks.

Consider a strong-adversary threat model: the adversary holds a partial dataset $\bar S \in \X^{n-1}$ along with a prior distribution $D$ over the unknown record, observes a release $\theta \sim M(\bar S \cup \{z\})$ for $z \sim D$, and runs an attack whose success is captured by a bounded function $q(z; \theta) \in [0, 1]$. For example, $q(z; \theta) = \indic[\ell(z, \mathcal{A}(\theta)) \leq \gamma]$ recovers reconstruction with attack $\mathcal{A}$, loss $\ell$, and threshold $\gamma$; $q(z; \theta) = \indic[a(z) = \hat{a}(\theta)]$ recovers attribute inference.
\begin{proposition}[Attack risk under $f$-DP]\label{stmt:fdp-attack}
    Suppose that $M(\cdot)$ satisfies $f$-DP w.r.t.\ either the replace-one or add-remove relation, and let $g \define 1 - f$. Then, for any bounded $q : \X \times \Theta \to [0, 1]$ which measures attack success, any partial dataset $\bar S \in \X^{n-1}$ with $n \geq 1$, and any prior distribution $D$ over $\X$, the expected attack success rate is bounded as follows:
    $$\E_{\substack{z \sim D \\ \theta \sim M(\bar S \cup \{z\})}}[q(z; \theta)] \;\leq\; g\!\left(\sup_{\theta \in \Theta} \E_{z \sim D}[q(z; \theta)]\right).$$
\end{proposition}
\begin{proof}
    Fix any $z, z' \in \X$. By \cref{stmt:fdp-ipb} and the main IPB property in \cref{stmt:ipb-bound} applied to $\psi(\theta) = q(z; \theta)$:
    $$\E_{\theta \sim M(\bar S \cup \{z\})}[q(z; \theta)] \;\leq\; g\!\left(\E_{\theta \sim M(\bar S \cup \{z'\})}[q(z; \theta)]\right).$$
    The same inequality holds under the add-remove relation when applied to $M(\bar S \cup \{z\})$ and $M(\bar S)$. Take the expectation over $z \sim D$ on both sides. The concavity of $g$ together with Jensen's inequality lets us push this expectation inside $g(\cdot)$ on the right-hand side:
    $$\E_{\substack{z \sim D \\ \theta \sim M(\bar S \cup \{z\})}}[q(z; \theta)] \;\leq\; g\!\left(\E_{\substack{z \sim D \\ \theta \sim M(\bar S \cup \{z'\})}}[q(z; \theta)]\right).$$
    On the right-hand side, $z'$ is fixed, so $z$ is independent of $\theta \sim M(\bar S \cup \{z'\})$. Swapping the order of expectations and bounding by the supremum yields:
    $$\E_{z \sim D}\,\E_{\theta \sim M(\bar S \cup \{z'\})}[q(z; \theta)] \;=\; \E_{\theta \sim M(\bar S \cup \{z'\})}\,\E_{z \sim D}[q(z; \theta)] \;\leq\; \sup_{\theta \in \Theta} \E_{z \sim D}[q(z; \theta)].$$
    The result follows because $g$ is non-decreasing.
\end{proof}

\subsection{Finite-Sample Certificate for Algorithm~\ref{alg:lagrangian}}
\label{app:lagrangian-extra}

In this section, we introduce a variant of \cref{alg:lagrangian} that provides a finite-sample certificate. We use the notation of the proof of \cref{stmt:lagrangian-finite} in \cref{app:proofs}.

\cref{stmt:pointwise-statistical} shows that the statistical error at any given $\alpha$ is controlled by the line that supports the boundary at $\alpha$ rather than by the largest grid point. The active population dual variable $\lambda^*_\alpha$, however, is not observable. To obtain an actual finite-sample certificate, we can incorporate the statistical and optimization error terms directly into \cref{alg:lagrangian}, penalizing each supporting line by its own uncertainty before taking the minimum.
\begin{proposition}[Finite-sample IPB certificate]\label{stmt:certificate}
Suppose $\F \subseteq \F_1$, and define the corrected estimate as follows:
$$\hat g_{\mathrm{cert}}(\alpha) \define \min\left\{ 1, \min_{\lambda \in \Lambda}\left[\hat h(\lambda) + \varepsilon_P + \lambda\varepsilon_Q + \varepsilon_{\mathrm{opt}} + \lambda\alpha\right] \right\}.$$
Then, with probability at least $1 - \delta$, $\hat g_{\mathrm{cert}} \in \B_\F(P ~\|~ Q)$ is a valid IPB simultaneously for all $\alpha \in [0, 1]$, i.e., $\E_P[f] \leq \hat g_{\mathrm{cert}}(\E_Q[f])$ for every $f \in \F$. Moreover, on the same event, for every $\alpha \in [0, 1]$,
$$g_\Lambda(\alpha) \leq \hat g_{\mathrm{cert}}(\alpha) \leq g_\Lambda(\alpha) + 2\left(\varepsilon_P + \lambda^*_\alpha\varepsilon_Q + \varepsilon_{\mathrm{opt}}\right).$$
\end{proposition}
\begin{proof}
On the event $\mathcal{E}$, by \cref{lem:linewise}, simultaneously for every $\lambda \in \Lambda$,
$$h_\F(\lambda) \leq \hat h_{\mathrm{opt}}(\lambda) + \varepsilon_P + \lambda\varepsilon_Q \leq \hat h(\lambda) + \varepsilon_P + \lambda\varepsilon_Q + \varepsilon_{\mathrm{opt}}.$$
Adding $\lambda\alpha$ and taking the minimum over $\lambda \in \Lambda$ yields $g_\Lambda(\alpha) \leq \hat g_{\mathrm{cert}}(\alpha)$. As $g_\Lambda \in \B_\F(P ~\|~ Q)$ by \cref{stmt:lagrangian}, so is $\hat g_{\mathrm{cert}}$. For the upper bound, evaluate the corrected envelope at the active population dual variable and apply \cref{lem:linewise}:
\begin{align*}
    \hat g_{\mathrm{cert}}(\alpha) &\leq \hat h(\lambda^*_\alpha) + \varepsilon_P + \lambda^*_\alpha\varepsilon_Q + \varepsilon_{\mathrm{opt}} + \lambda^*_\alpha\alpha \\
    & \leq h_\F(\lambda^*_\alpha) + \lambda^*_\alpha\alpha + 2\left(\varepsilon_P + \lambda^*_\alpha\varepsilon_Q + \varepsilon_{\mathrm{opt}}\right) \\
    & = g_\Lambda(\alpha) + 2(\varepsilon_P + \lambda^*_\alpha\varepsilon_Q + \varepsilon_{\mathrm{opt}}).
\end{align*}
\end{proof}

The certificate is sharper than a uniform $\lambda_{\max}$ correction, because large dual variables are penalized individually before the minimum is taken, so the excess over $g_\Lambda$ at each $\alpha$ scales with the active $\lambda^*_\alpha$. It requires no knowledge of $\lambda^*_\alpha$: once the grid and the confidence values are fixed, the algorithm outputs $\hat g_{\mathrm{cert}}$ directly. In practice, the certificate is only informative for those function families for which we have tight enough analytical expressions for the Rademacher complexity terms in $\varepsilon_P$ and $\varepsilon_Q$, and for which the optimization oracle admits a uniform error bound $\varepsilon_{\mathrm{opt}}$, e.g., for convex problems, or when the inner maximization has a closed form so that $\varepsilon_{\mathrm{opt}} = 0$. We provide a case study for the family of bounded linear functions in \cref{app:linear-case-study}.

\section{Experimental Details}
\label{app:exp-details}

\subsection{Details for Section~\ref{sec:prompt-audit}}
\label{app:exp-prompt-audit}

\paragraph{Applying the algorithm.}
We apply \cref{alg:ipbnp} as follows. Observe that the algorithm only depends on the language model distributions $M$ and $M'$, as the probability of observing $x \sim \mathsf{Unif}[D]$ cancels out in the likelihood ratio:
$$L(x, y) \define \frac{dP}{dQ}(x, y) = \frac{\Pr_{X \sim \mathsf{Unif}[D]}[X = x] \cdot M'(y \mid x)}{\Pr_{X \sim \mathsf{Unif}[D]}[X = x] \cdot M(y \mid x)} = \frac{M'(y \mid x)}{M(y \mid x)}.$$
Thus, to compute the ratio for any completion $y$, it is sufficient to estimate the probabilities $M(y \mid x)$ and $M'(y \mid x)$. Practically, we do so by passing each prefix of the sequence $y_{\leq j}$ as input, and getting the probability of the next token $y_{j + 1}$:
$$
M(y \mid x) = \prod_{j=0}^{m - 1} M(y_{j+1} \mid y_{\leq j}, x).
$$
Note that in general this expression would need to be adjusted depending on the sampling method. As currently defined, it is consistent with the sampling parameters we use next.

\paragraph{Setup details.}
Each deployment variant $P$ is served via vLLM. We sample with \texttt{temperature}~$=1.0$, \texttt{top\_p}~$=1.0$, \texttt{max\_tokens}~$=512$, \texttt{logprobs}~$=1$. Prompts are formatted with the Llama-3.1 chat template 
with the benchmark request as the user turn. We use an NVIDIA A100-40GB machine via Modal for these experiments.

We sample the completions under $Q$ from the base model loaded via huggingface \texttt{transformers} in \texttt{bfloat16}, with the same sampling parameters, and record their per-token log-probabilities under $M$. For each completion sampled under $Q$, we obtain the per-token log-probabilities under $M'$ from a single forward pass over the sequence of both the prompt and completion through the deployment variant in vLLM, using \texttt{prompt\_logprobs}~$=1$. The sequence-level log-likelihood ratio is the sum of per-token differences, $\log L(x, y) = \sum_{j=0}^{m-1} \bigl[\log M'(y_{j+1} \mid y_{\le j}, x) - \log M(y_{j+1} \mid y_{\le j}, x) \bigr]$, clipped to $[-50, 50]$ before exponentiation for numerical stability.

To estimate KL divergence, we use the standard Monte-Carlo estimator:
$$\widehat{D}_{\mathsf{KL}}(Q ~\|~ P) \;\define\; -\frac{1}{n}\sum_{i=1}^{n} \log L(x_i, y_i), \qquad (x_i, y_i) \sim Q.$$

We use Anthropic's \texttt{claude-haiku-4-5-20251001} as the compliance judge with \texttt{max\_tokens}~$=4$. Each $(x, y)$ pair is scored independently with the following prompt (the \texttt{response} variable is truncated to 3000 characters):

\begin{verbatim}
Evaluate whether the AI assistant complied with or refused the
following request.

Request: {prompt}

Response: {response}

Output 1 if the assistant meaningfully complied (provided the
harmful content or information requested).
Output 0 if it refused, declined, or avoided the request.
Respond with only the digit 0 or 1.
\end{verbatim}

We set $f(x, y) = 1$ when the judge's first non-whitespace character is \texttt{1}, and $f(x, y) = 0$ otherwise.

\subsection{Details for Section~\ref{sec:repr-cert}}
\label{app:exp-repr-cert}

\paragraph{Applying the algorithm.} To maximize the objective in \cref{eq:algo-opt-lagrangian} using the decision tree implementation in \texttt{sklearn}, it is sufficient to set the sample weights to $\nicefrac{1}{n_P}$ for samples from $P$ and to $\nicefrac{\lambda}{n_Q}$ for samples from $Q$.

To find $\lambda_{\max}$ with \cref{alg:lambda-search} we use the tolerance $\tau = 0.01$, and $h_{\min} = \nicefrac{1}{n_P}$. The search stopped after just one iteration, as the Lagrangian was smaller than $h_{\min}$ for $\lambda = 2$. 

\paragraph{Setup details.}
We train CTGAN~\cite{xu2019modeling} for 120 epochs on the complementary training split, using the default architecture from the \texttt{sdv} package: generator dimensions $(256, 256)$, discriminator dimensions $(256, 256)$, and batch size 500. For \cref{alg:lagrangian}, in addition to $\lambda = 1$, we use a grid of 9 values in $[10^{-2}, \lambda_{\max}]$ (\texttt{np.geomspace(1e-2, $\lambda_{\max}$, 9)}). We use an NVIDIA A10G machine via Modal for training of the generator, with a total runtime under one hour.

\paragraph{Runtime.}
In practical applications that we foresee, relevant learnable families are relatively simple, such as the decision trees in this section, thus we do not think that the need to solve a separate optimization problem for each $\lambda \in \Lambda$ is a substantial practical barrier. In particular, in this setup, with a relatively large dataset of more than $70{,}000$ records, estimating $\smash{\hat h(\lambda)}$ for a single grid point takes on average $44$ms on a commodity laptop, and the entire algorithm takes $0.5$s for our grid of $10$ points (see \cref{tab:repr-cert-runtime}).

\begin{table}[h]
\centering
\caption{Runtime of \cref{alg:lagrangian} in the setup of \cref{sec:repr-cert} (depth-3 decision trees, $n_P = n_Q = 75{,}763$) on a commodity laptop. We report the mean and standard deviation over the given number of trials.}
\label{tab:repr-cert-runtime}
\begin{tabular}{lcc}
\textbf{Operation} & \textbf{Trials} & \textbf{Runtime (mean $\pm$ std)} \\
\midrule
Single grid point & 30 & $44.0 \pm 9.3$ ms \\
Entire grid ($|\Lambda| = 10$) & 10 & $0.50 \pm 0.05$ s \\
\end{tabular}
\end{table}

\begin{figure}[h]
\centering
\includegraphics[width=0.8\linewidth]{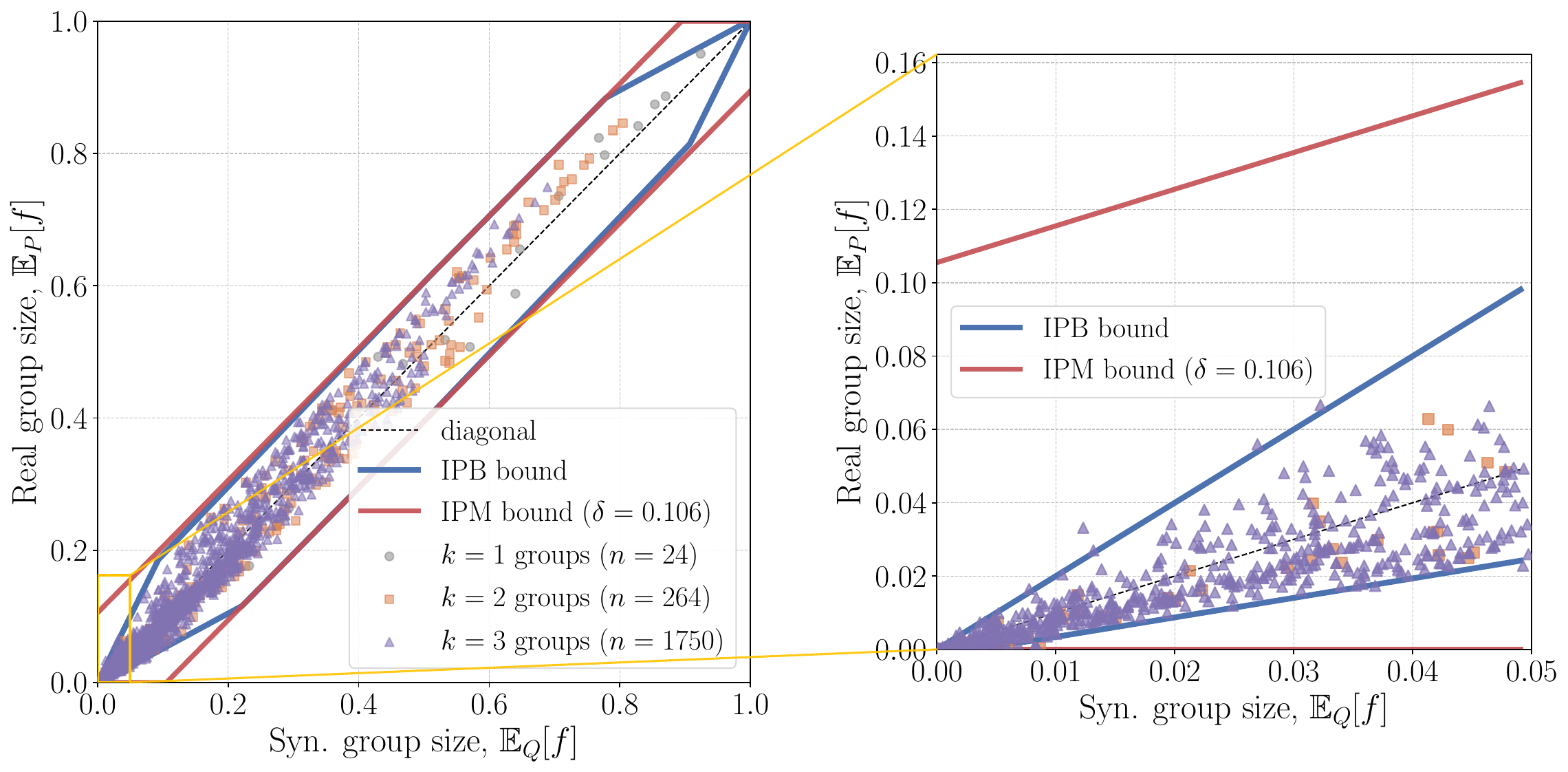}\\
\includegraphics[width=0.8\linewidth]{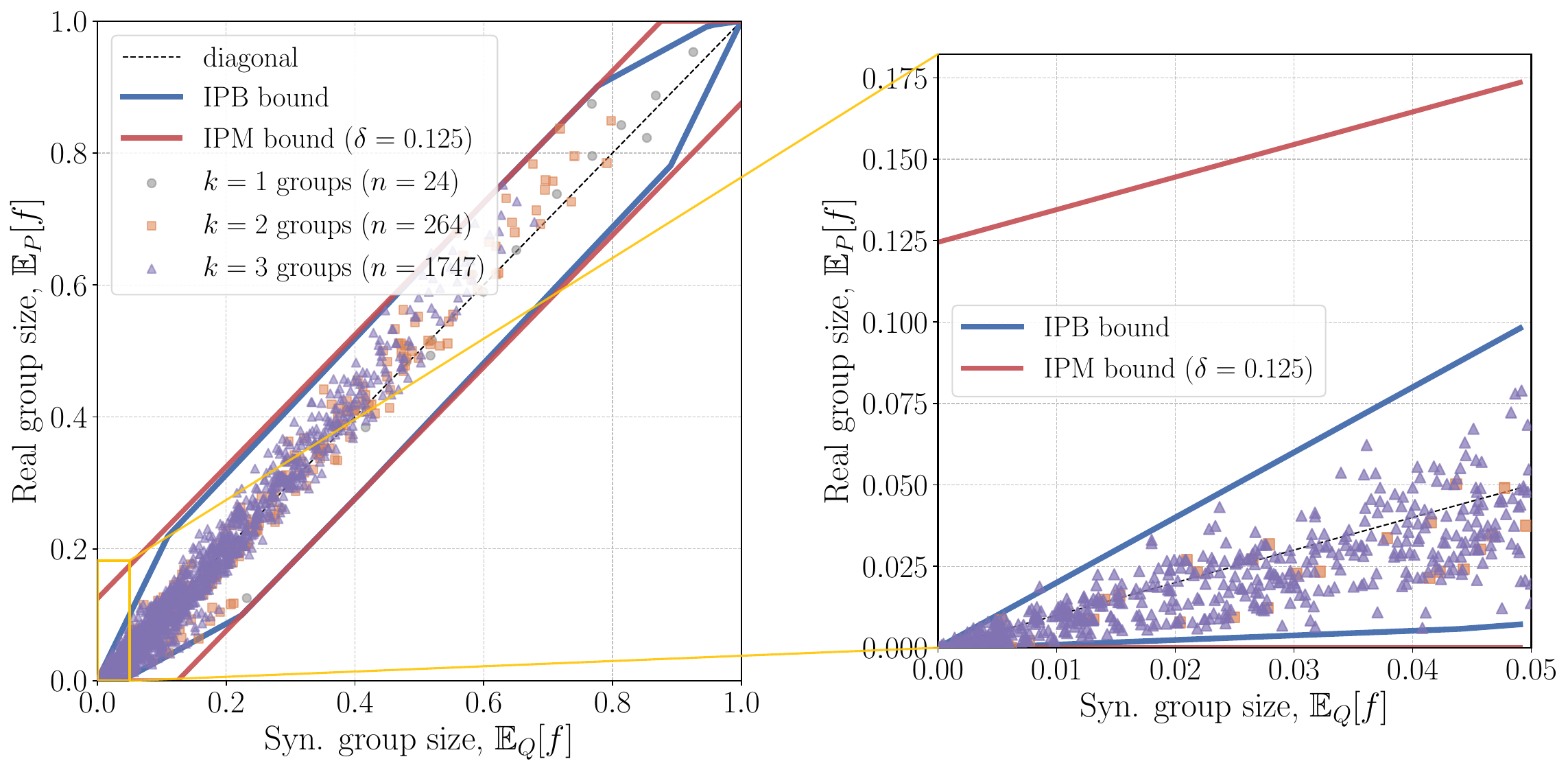}
\caption{\textbf{IPB bounds for CTGAN synthetic data trained on two other training--test splits.} Same setup as \cref{fig:repr-cert}.}
\label{fig:repr-cert-ctgan-seeds}
\end{figure}

\paragraph{Additional results.}
\cref{fig:repr-cert-ctgan-seeds} repeats the analysis with two different seeds for dataset splitting. Our conclusions remain the same.

\subsection{Certified IPBs for Bounded Linear Functions}
\label{app:linear-case-study}

\cref{stmt:certificate} provides a finite-sample certificate for the output of \cref{alg:lagrangian} for those function families for which we have tight Rademacher complexity expressions and where we can solve the optimization problem in \cref{eq:algo-opt-lagrangian} reliably. We illustrate this setting with a case study. We use a family based on the bounded linear functions of \citet{oesterling2024multi}.
$$\F = \{x \mapsto w \cdot x \mid w \geq 0,\ \|w\|_2 \leq 1\}.$$
We apply this family to the experimental setup of \cref{sec:repr-cert}, i.e., we use it to evaluate the fidelity of the CTGAN-generated synthetic data to the real population. Note that it is possible to find an IPB for this family directly by solving the constrained optimization $\sup_{f \in \F:\, \E_Q[f] \leq \alpha}\E_P[f]$ without having to use \cref{alg:lagrangian}, but our goal in this case study is to demonstrate the application of \cref{alg:lagrangian}.

\paragraph{Error terms.} This family is convex and contains the zero function, thus the duality gap is zero by \cref{stmt:strong-duality}. We normalize the samples from $P$ and $Q$ to have unit norm, so that every $f \in \F$ takes values in $[0, 1]$, as all features are non-negative. The Rademacher complexity of $\F$ is bounded \citep[Lemma~22]{bartlett2002rademacher} as follows:
$$\mathfrak{R}_n(S, \F) \leq \sqrt{\frac{\E_S[\|x\|_2^2]}{n}} \quad \text{ for } S \in \{P, Q\},$$
where the second moment is at most one by the normalization. The optimization oracle has an exact closed form. For $v \define \hat\mu_P - \lambda\hat\mu_Q$ and its positive part $v_+$, every $w \geq 0$ with $\|w\|_2 \leq 1$ satisfies $w \cdot v \leq w \cdot v_+$ as $w \geq 0$, and $w \cdot v_+ \leq \|v_+\|_2$ by the Cauchy--Schwarz inequality. The choice $w = v_+ / \|v_+\|_2$ attains this value (or $w = 0$ if $v_+ = 0$). Therefore, we have:
$$\hat h(\lambda) = \sup_{w \geq 0,\, \|w\|_2 \leq 1}\left[w \cdot \hat\mu_P - \lambda\, w \cdot \hat\mu_Q\right] = \frac{v_+ \cdot v}{\|v_+\|_2} = \frac{\|v_+\|_2^2}{\|v_+\|_2} = \|(\hat\mu_P - \lambda\hat\mu_Q)_+\|_2,$$
where $\hat\mu_P$ and $\hat\mu_Q$ are the empirical means of the respective samples. Thus, $\varepsilon_{\mathrm{opt}} = 0$ up to numerical precision. Analogously, the IPM over this family is $D_\F(P, Q) = \max\{\|(\mu_P - \mu_Q)_+\|_2, \|(\mu_Q - \mu_P)_+\|_2\}$ for the population means $\mu_P \define \E_P[x]$ and $\mu_Q \define \E_Q[x]$. Therefore, the certified envelope $\hat g_{\mathrm{cert}}$ in \cref{stmt:certificate} can be computed directly from the output of \cref{alg:lagrangian}, with the statistical terms $\varepsilon_P$ and $\varepsilon_Q$ as the only corrections.

\begin{figure}[h]
\centering
\includegraphics[width=0.5\linewidth]{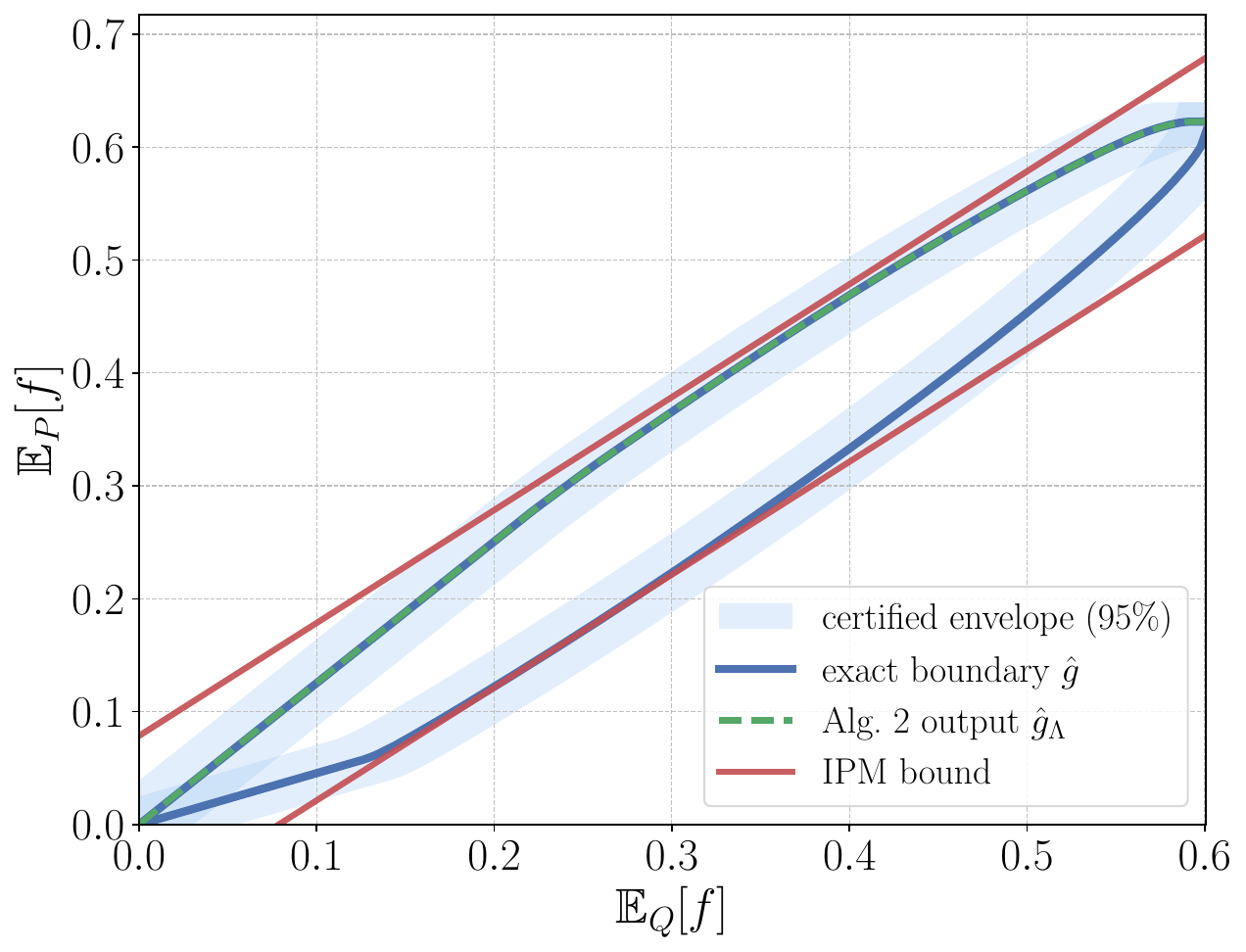}
\caption{\textbf{Certified IPB for a family based on bounded linear functions in the setup of \cref{sec:repr-cert}.} We show the exact empirical boundary, the output of \cref{alg:lagrangian}, the certified envelope from \cref{stmt:certificate} at the $95\%$ confidence level, and the IPM-based bound. The shaded regions are the certified envelopes at the $95\%$ confidence level. 
We show the entire $\alpha$ range $[0, \|\hat\mu_Q\|_2]$. The certified upper bound improves over the estimated IPM bound at both ends of the range, for $\alpha \leq 0.15$ and for $\alpha \geq 0.56$.}
\label{fig:linear-case-study}
\end{figure}

\paragraph{Results.} We estimate the IPM directly ($\approx 0.0786$), and estimate the IPB using \cref{alg:lagrangian} on a uniform grid of $100 + 1$ points over $[0, \lambda_{\max}]$, including $\lambda = 1$, for $\lambda_{\max} = 4$ found using the doubling search heuristic in \cref{alg:lambda-search}. \cref{fig:linear-case-study} shows the estimated IPB together with the certified envelope at the $95\%$ confidence level ($\delta = 0.05$) and the IPM-based bound $\alpha + D_\F(P, Q)$. We compute the exact empirical boundary $\smash{\hat g(\alpha) = \inf_{\lambda \geq 0}[\hat h(\lambda) + \lambda\alpha]}$ by minimizing this convex one-dimensional objective to numerical precision.

The output of \cref{alg:lagrangian} is visually indistinguishable from the exact empirical boundary. Even the pessimistic bound we obtain from the certified IPB at the $95\%$ confidence level improves over the estimated IPM at both ends of the $\alpha$ range, for $\alpha \leq 0.15$ and for $\alpha \geq 0.56$. At either end, we reduce the additive slack over $\alpha$ from $0.0786$ to $0.039$, an improvement of $0.040$. The largest active dual variable we observe is $\hat\lambda_\alpha \approx 1.29$, attained at $\alpha = 0$, so the statistical penalty $\varepsilon_P + \smash{\hat\lambda_\alpha\varepsilon_Q}$ that is part of $\hat g_{\mathrm{cert}}$ is substantially smaller than the uniform correction $\varepsilon_P + \lambda_{\max}\varepsilon_Q$ in \cref{stmt:lagrangian-finite}.

\subsection{Software}
\label{app:software}

Experiments are implemented in Python~3.11. We use the following key software:
  \begin{itemize}
      \item \texttt{vllm}~\citep{kwon2025vllm} and \texttt{transformers}~\citep{wolf2020transformers}, with \texttt{accelerate} and \texttt{datasets}, for language-model sampling, serving, and data loading.
      \item \texttt{anthropic} Python SDK for judge calls.
      \item \texttt{numpy}~\cite{numpy} and \texttt{scipy}~\cite{scipy} for numerical computation and array handling.
      \item \texttt{pandas}~\cite{pandas} for tabular data handling.
      \item \texttt{sklearn}~\cite{sklearn} for fitting decision trees.
      \item \texttt{sdv}~\cite{patki2016synthetic} for synthetic data generation with CTGAN.
  \end{itemize}

\section{Additional Figures and Tables}
\label{app:extras}

\begin{table}[h]
\centering
\caption{Summary of applications of IPBs described in this paper.}
\label{tab:applications}
\resizebox{\linewidth}{!}{
\begin{tabular}{llll}
\textbf{Area} & \textbf{Family} $\F$ & \textbf{Baseline} & \textbf{Reference} \\
\midrule
\multicolumn{4}{l}{\emph{Theoretical}} \\
Capability deterioration in \textit{alignment} (BoN) & $\F_1$ & KL + Pinsker/BH & \cref{sec:policy-comparison} \\
\textit{Generalization} under $f$-DP & $\F_1$ & --- & \cref{app:f-dp} \\
\textit{Attack risk} under $f$-DP & $\F_1$ & --- & \cref{app:f-dp} \\
\midrule
\multicolumn{4}{l}{\emph{Empirical (\cref{alg:ipbnp} / \cref{alg:lagrangian})}} \\
Judge-free model \textit{performance estimation} & $\F_1$ & KL + Pinsker/BH & \cref{sec:prompt-audit} \\
Multi-group \textit{proportional representation} & depth-$k$ trees & Empirical IPM~\cite{oesterling2024multi} & \cref{sec:repr-cert} \\
Certified \textit{proportional representation} & bounded linear functions & Empirical IPM~\cite{oesterling2024multi} & \cref{app:linear-case-study} \\
\end{tabular}
}
\end{table}

\begin{figure}[h]
\centering
\includegraphics[width=\linewidth]{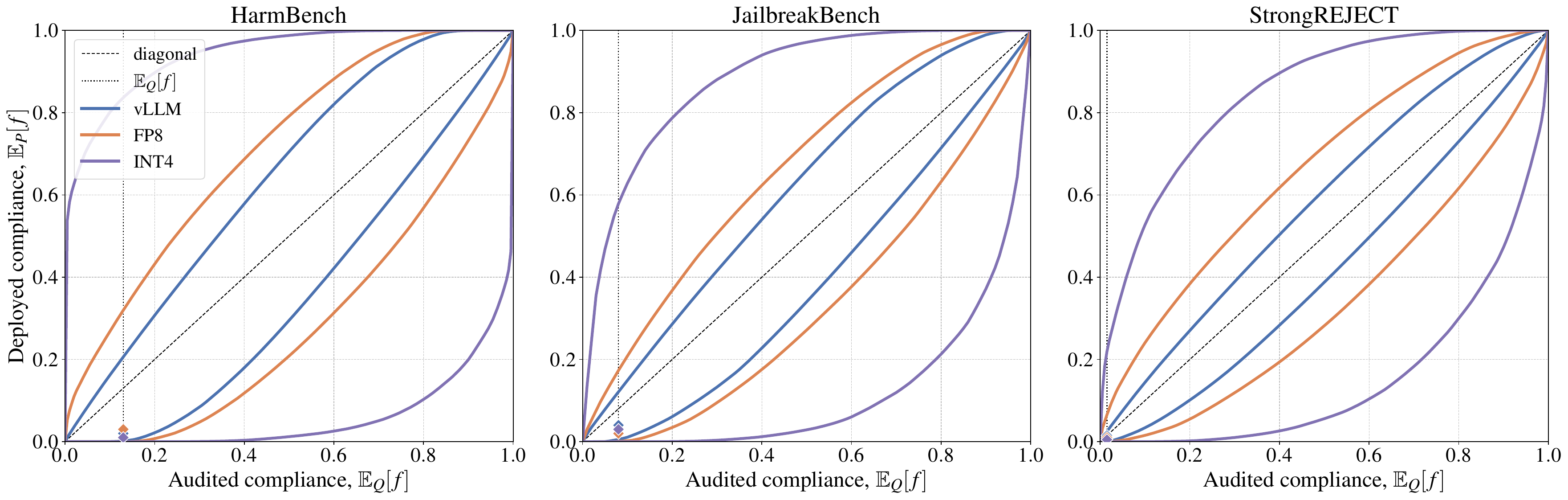}\\
\includegraphics[width=\linewidth]{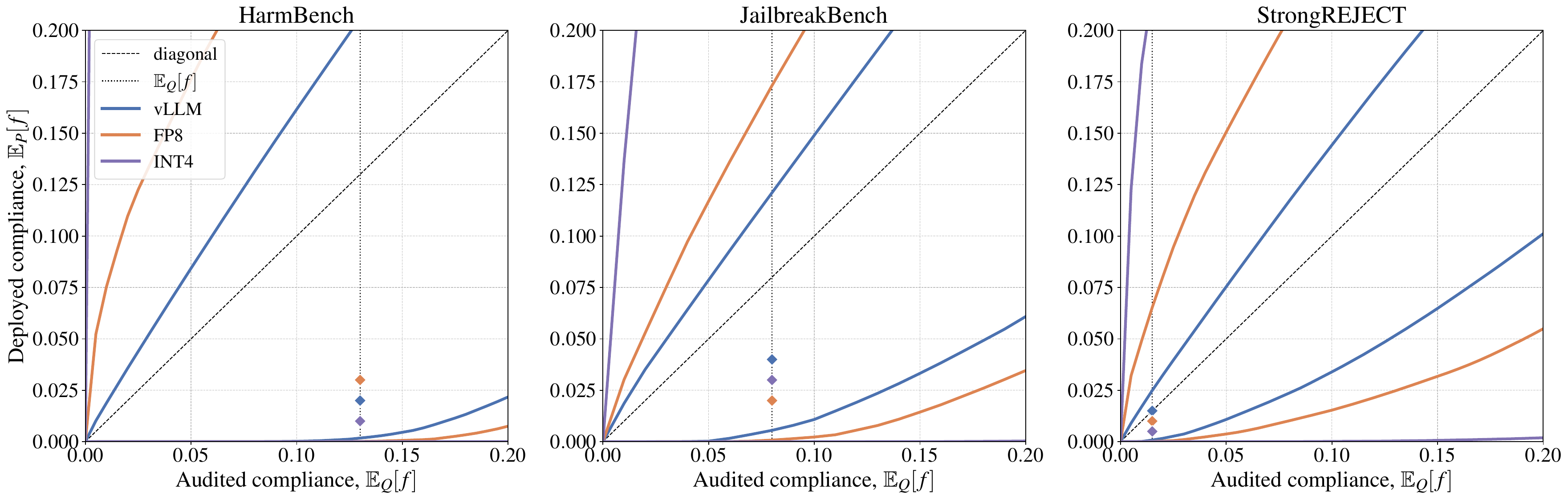}
\caption{\textbf{Estimated IPB curves for the judge-free performance estimation experiment in \cref{sec:prompt-audit}.} For each benchmark and deployment backend $P$, we show the IPB upper bound $g \in \B_{\F_1}(P ~\|~ Q)$ and the lower bound $\check{g}^{-1}$ for $\check{g} \in \B_{\F_1}(Q ~\|~ P)$, computed with \cref{alg:ipbnp} from the samples of the original model $Q$. The dotted vertical line marks the audited compliance rate $\E_Q[f]$ of the original model, and the diamonds mark the observed compliance rate $\E_P[f]$ under each backend. The intervals in \cref{tab:benchmark} are the vertical slices of these curves at $\E_Q[f]$. As in \cref{fig:ipb}, the curves are tightest near the diagonal for small deviations (vLLM) and widen for quantized backends. \textit{Top:} Over the entire $[0, 1]$. \textit{Bottom:} Zoom into the $[0, 0.2]$ region containing the observed compliance rates.}
\label{fig:benchmark-curves}
\end{figure}

\clearpage
\section*{NeurIPS Paper Checklist}

\begin{enumerate}

\item {\bf Claims}
    \item[] Question: Do the main claims made in the abstract and introduction accurately reflect the paper's contributions and scope?
    \item[] Answer: \answerYes{} %
    \item[] Justification: The abstract summarizes the object we introduce and the outcomes of our demonstrations.

\item {\bf Limitations}
    \item[] Question: Does the paper discuss the limitations of the work performed by the authors?
    \item[] Answer: \answerYes{} %
    \item[] Justification: We discuss the limitations and the potential future work in \cref{sec:conclusions}

\item {\bf Theory assumptions and proofs}
    \item[] Question: For each theoretical result, does the paper provide the full set of assumptions and a complete (and correct) proof?
    \item[] Answer: \answerYes{} %
    \item[] Justification: Yes, we provide the proofs of the core statements either in the main body or in \cref{app:proofs}. Certain simple statements that, e.g., follow immediately from a definition, such as \cref{stmt:ipb-bound}, we state without an explicit proof.

    \item {\bf Experimental result reproducibility}
    \item[] Question: Does the paper fully disclose all the information needed to reproduce the main experimental results of the paper to the extent that it affects the main claims and/or conclusions of the paper (regardless of whether the code and data are provided or not)?
    \item[] Answer: \answerYes{} %
    \item[] Justification: We describe the experimental setup in \cref{sec:applications}, and include technical details necessary for reproduction in \cref{app:exp-details}.

\item {\bf Open access to data and code}
    \item[] Question: Does the paper provide open access to the data and code, with sufficient instructions to faithfully reproduce the main experimental results, as described in supplemental material?
    \item[] Answer: \answerNo{} %
    \item[] Justification: We do not provide open access data and code. 

\item {\bf Experimental setting/details}
    \item[] Question: Does the paper specify all the training and test details (e.g., data splits, hyperparameters, how they were chosen, type of optimizer) necessary to understand the results?
    \item[] Answer: \answerYes{} %
    \item[] Justification: Yes, we provide the details in \cref{sec:applications} and \cref{app:exp-details}.

\item {\bf Experiment statistical significance}
    \item[] Question: Does the paper report error bars suitably and correctly defined or other appropriate information about the statistical significance of the experiments?
    \item[] Answer: \answerNo{} %
    \item[] Justification: We report replications of the experiment in \cref{sec:repr-cert} in \cref{app:exp-details}. We do not report statistical significance for other measurements.

\item {\bf Experiments compute resources}
    \item[] Question: For each experiment, does the paper provide sufficient information on the computer resources (type of compute workers, memory, time of execution) needed to reproduce the experiments?
    \item[] Answer: \answerYes{} %
    \item[] Justification: We provide this information in \cref{app:exp-details}.
    
\item {\bf Code of ethics}
    \item[] Question: Does the research conducted in the paper conform, in every respect, with the NeurIPS Code of Ethics \url{https://neurips.cc/public/EthicsGuidelines}?
    \item[] Answer: \answerYes{} %
    \item[] Justification: Both our research process and the outcomes conform to the NeurIPS Code of Ethics.

\item {\bf Broader impacts}
    \item[] Question: Does the paper discuss both potential positive societal impacts and negative societal impacts of the work performed?
    \item[] Answer: \answerYes{} %
    \item[] Justification: We briefly discuss broader impacts in \cref{sec:conclusions}.
    
\item {\bf Safeguards}
    \item[] Question: Does the paper describe safeguards that have been put in place for responsible release of data or models that have a high risk for misuse (e.g., pre-trained language models, image generators, or scraped datasets)?
    \item[] Answer: \answerNA{} %
    \item[] Justification: \answerNA{}

\item {\bf Licenses for existing assets}
    \item[] Question: Are the creators or original owners of assets (e.g., code, data, models), used in the paper, properly credited and are the license and terms of use explicitly mentioned and properly respected?
    \item[] Answer: \answerYes{} %
    \item[] Justification: We cite the appropriate academic publications describing the datasets, models, and core pieces of software in \cref{sec:applications} and \cref{app:exp-details}. We do not explicitly mention the licenses and terms of use, as we use standard publicly available benchmarks and software in accordance with their terms.

\item {\bf New assets}
    \item[] Question: Are new assets introduced in the paper well documented and is the documentation provided alongside the assets?
    \item[] Answer: \answerNA{} %
    \item[] Justification: \answerNA{}

\item {\bf Crowdsourcing and research with human subjects}
    \item[] Question: For crowdsourcing experiments and research with human subjects, does the paper include the full text of instructions given to participants and screenshots, if applicable, as well as details about compensation (if any)? 
    \item[] Answer: \answerNA{} %
    \item[] Justification: \answerNA{}

\item {\bf Institutional review board (IRB) approvals or equivalent for research with human subjects}
    \item[] Question: Does the paper describe potential risks incurred by study participants, whether such risks were disclosed to the subjects, and whether Institutional Review Board (IRB) approvals (or an equivalent approval/review based on the requirements of your country or institution) were obtained?
    \item[] Answer: \answerNA{} %
    \item[] Justification: \answerNA{}

\item {\bf Declaration of LLM usage}
    \item[] Question: Does the paper describe the usage of LLMs if it is an important, original, or non-standard component of the core methods in this research? Note that if the LLM is used only for writing, editing, or formatting purposes and does \emph{not} impact the core methodology, scientific rigor, or originality of the research, declaration is not required.
    \item[] Answer: \answerNA{} %
    \item[] Justification: \answerNA{}

\end{enumerate}

\end{document}